\pdfoutput=1
\documentclass[11pt]{article}

\usepackage[final]{acl}

\usepackage{times}
\usepackage{latexsym}
\usepackage[T1]{fontenc}
\usepackage[utf8]{inputenc}
\usepackage{microtype}
\usepackage{inconsolata}
\usepackage{graphicx,enumitem,subfig,amsmath,amssymb,tikz,xcolor,colortbl,arydshln,multirow}

\usepackage{xspace}
\usepackage{mdwlist}
\usepackage{tabularx}
\usepackage{booktabs}
\usepackage{colortbl}

\newcommand{\rparagraph}[1]{\vspace{0.0mm}\noindent\textbf{#1.}}

\newcommand{\sparagraphnodot}[1]{\vspace{0.0mm}\noindent\textbf{#1}}

\newcommand{\colap}{\texttt{CoLAP}\xspace}
\newcommand{\xrcl}{$\mathit{XRCL}$\xspace}
\newcommand{\xrclmath}{\mathit{XRCL}\xspace}
\newcommand{\xccl}{$\mathit{XCCL}$\xspace}
\newcommand{\xcclmath}{\mathit{XCCL}\xspace}
\newcommand{\eos}{\texttt{<EOS>}\xspace}

\definecolor{darkgreen}{HTML}{364e00}
\definecolor{steelblue}{HTML}{4682B4}
\definecolor{lightblue}{RGB}{220, 230, 241}

\newcolumntype{Y}{>{\centering\arraybackslash}X}

\title{Bridging Language Gaps: Enhancing Few-Shot Language Adaptation}

\author{
	Philipp Borchert$^{1,2}$, Jochen De Weerdt$^2$, Marie-Francine Moens$^3$\\
	\textsuperscript{1}IESEG School of Management, 3 Rue de la Digue, 59000 Lille, France\\
	\textsuperscript{2}Research Centre for Information Systems Engineering, KU Leuven, Belgium\\
	\textsuperscript{3}Department of Computer Science, KU Leuven, Belgium\\
}

\begin{document}
\maketitle
\begin{abstract}

The disparity in language resources poses a challenge in multilingual NLP, with high-resource languages benefiting from extensive data, while low-resource languages lack sufficient data for effective training. Our \textbf{Co}ntrastive \textbf{L}anguage \textbf{A}lignment with \textbf{P}rompting (\colap) method addresses this gap by integrating contrastive learning with cross-lingual representations, facilitating task-specific knowledge transfer from high-resource to lower-resource languages. The primary advantage of our approach is its data efficiency, enabling rapid adaptation to new languages and reducing the need for large labeled datasets.
We conduct experiments with multilingual encoder-only and decoder-only language models on natural language understanding tasks, including natural language inference and relation extraction, evaluating performance across both high- and low-resource languages. Our results demonstrate that \colap outperforms few-shot cross-lingual transfer baselines and in-context learning, even with limited available data. This effectively narrows the cross-lingual performance gap, contributing to the development of more efficient multilingual NLP techniques.\footnote{Code: \scriptsize{\url{https://github.com/pnborchert/CoLAP}.}}
% https://anonymous.4open.science/r/CoLAP-B2BE

\end{abstract}

\section{Introduction}
\label{sec:introduction}

The adaptation of pretrained language models (PLMs) to specific downstream tasks is typically resource-intensive, requiring extensive labeled data and computational resources. This becomes particularly challenging in multilingual contexts, especially for languages not represented in the model's pretraining data. These low-resource languages are significantly underrepresented, primarily due to the scarcity of large-scale corpora necessary for self-supervised pretraining and the lack of labeled data required for task-specific fine-tuning \citep{lauscher-etal-2020-zero, wu-dredze-2020-languages, yang-etal-2022-enhancing}. Multilingual PLMs, such as mBERT \citep{devlin-etal-2019-bert}, XLM-R \citep{conneau-etal-2020-unsupervised}, and Mistral \citep{jiang2023mistral7b}, have been developed to address these challenges. These models are trained across a wide range of languages using self-supervised data within a unified embedding space, facilitating cross-lingual transfer (XLT) of information. They have shown remarkable results in multilingual tasks. Despite their broad versatility, multilingual PLMs often exhibit representation degradation, particularly in low-resource languages \citep{yang-etal-2022-enhancing}. This degradation stems from the skewed distribution of self-supervised training data, which disproportionately favors high-resource languages. As a result, the representational quality for low-resource languages, and particularly for those not included in the pretraining data, is substantially diminished, leading to a decline in downstream task performance \citep{winata-etal-2022-cross}.

\begin{figure*}
\vspace{-10pt}
\centering
    \includegraphics[width=0.9\linewidth]{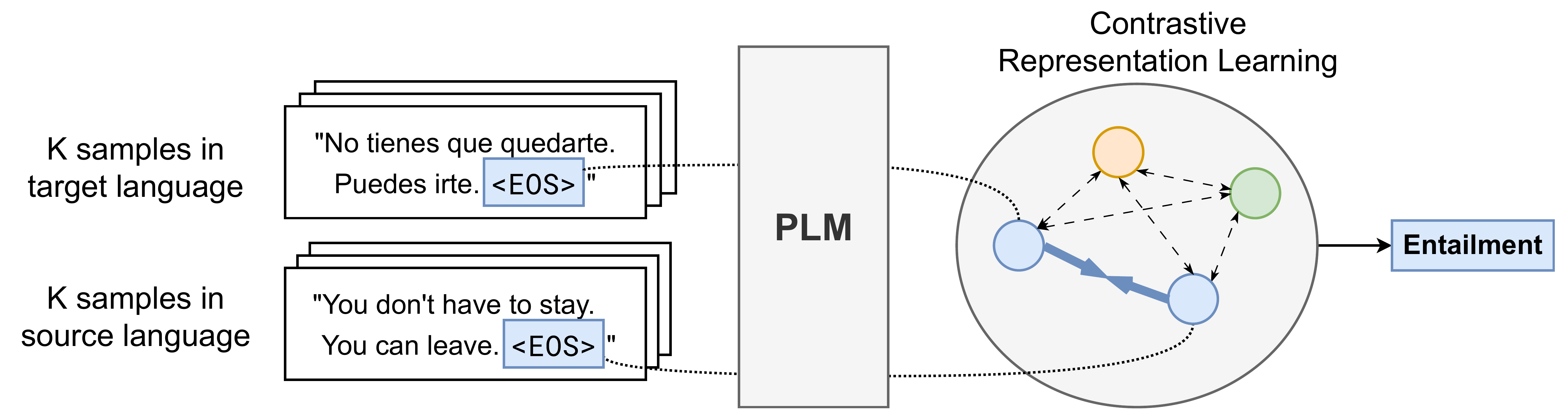}
    \caption{Illustration of our contrastive language alignment with prompting (\colap) approach for few-shot cross-lingual transfer applied to the natural language inference task.}
    \label{fig:model}
    \vspace{-10pt}
\end{figure*}

To enhance model representations and downstream task performance, models are commonly fine-tuned using task-specific data. However, acquiring such labeled data at scale is a significant challenge, especially for languages already disadvantaged by insufficient pretraining data \citep{winata-etal-2022-cross}. While multilingual PLMs have advanced performance for these languages compared to monolingual models primarily trained on English data, XLT remains challenging as there are substantial performance disparities between different languages. This disparity becomes more evident when fine-tuning PLMs on task data using high-resource languages and subsequently evaluating the fine-tuned model on other languages \citep{guo-etal-2023-analyzing}. 
% We display the cross-lingual transfer gap visually for two of the datasets investigated in this study in Figure \ref{fig:xlt_gap}.
Few-shot learning techniques address this gap, enabling models to learn effectively from limited data. Recent works in this domain encompass a range of zero-shot cross-lingual transfer (ZS-XLT) and few-shot cross-lingual transfer (FS-XLT) methods, including full fine-tuning \citep{lauscher-etal-2020-zero}, in-context learning \citep{winata-etal-2022-cross}, prompting \citep{schick-schutze-2021-exploiting}, representation mixup \citep{yang-etal-2022-enhancing,xu-etal-2023-language-representation}, and contrastive learning \citep{chi-etal-2021-infoxlm}. Among these, prompting techniques have shown efficacy in low-resource settings. Prompts express tasks as language modeling problems without introducing new model parameters, often articulated as natural language sentences \citep{qi-etal-2022-enhancing, nie-etal-2023-cross}. Despite these advances, full fine-tuning remains standard practice in the field \citep{zhao-schutze-2021-discrete,zhou-etal-2023-enhancing}.

In our study, we propose contrastive language alignment with prompting (\colap), addressing the cross-lingual transfer gap by aligning representations of underrepresented languages with those from high-resource languages, especially English. Instead of aligning representations during language modeling, we focus on the application of pretrained models on specific downstream tasks. We hypothesize that the transfer of discriminative task-specific information from English to lower-resource languages can be achieved efficiently without relying on abundant self-supervised or task-specific labeled data. This builds upon prior research by \citet{chi-etal-2021-infoxlm} and \citet{yang-etal-2022-enhancing}, which demonstrated the effectiveness of such alignment in general domain representations and ZS-XLT. 
We posit that representations for downstream classification tasks are less complex than representations used for language modeling because they only capture information relevant to the downstream objective. As a result, these representations can be transferred more data-efficiently between languages. We evaluate a representation contrastive learning approach (\xrcl) that aligns source and target language representations based on parallel translations. Additionally, we introduce a class contrastive learning objective (\xccl), which aligns representations of instances sharing the same class labels across languages, which does not require parallel translations. This facilitates the data annotation process and reduces costs.

Our experiments focus on natural language understanding tasks, including natural language inference and relation extraction, across 27 languages. We evaluate multilingual encoder-only and decoder-only models in the FS-XLT setting, where models are fine-tuned in a high-resource source language and subsequently adapted to the target language using few-shot learning. Our results demonstrate that \colap improves upon strong FS-XLT benchmarks but also outperforms in-context learning, even with limited available data. This methodology aims to narrow the cross-lingual performance gap, thereby extending the benefits of NLP models to a broader range of languages.

\rparagraph{Contributions}
    \textbf{1)} We introduce \colap, a novel few-shot cross-lingual transfer method that leverages contrastive learning to efficiently transfer knowledge from high- to lower-resource languages.
    \textbf{2)} \colap enhances cross-lingual transfer performance for low-resource languages, including those not represented in PLM pretraining.
    \textbf{3)} We propose \xccl, a contrastive learning objective that enhances FS-XLT performance without relying on parallel translations.
    \textbf{4)} We propose a few-shot exemplar selection approach based on representation similarity improving data efficiency.

\section{Related Work}
\label{sec:related_work}

\textbf{Cross-lingual transfer} involves transferring knowledge from a source language to one or more target languages \citep{ruder_2019_survey}. It is grounded in pretraining language models on multilingual data, embedding languages within a shared universal space to facilitate cross-lingual information transfer. This approach is particularly effective for languages included in the pretraining dataset \citep{ruder_2019_survey, ansell-etal-2021-mad-g}. Multilingual PLMs have extended this capability, showing promise in ZS-XLT even for languages not directly seen during pretraining \citep{devlin-etal-2019-bert, conneau-etal-2020-unsupervised}. The transfer's efficacy is influenced by the alignment quality of representations between source and target languages \citep{wu-dredze-2020-languages, cao_2020_multilingual}, which is especially effective for topologically similar languages with large-scale pretraining corpora \citep{lauscher-etal-2020-zero}. Recent advancements in enhancing XLT performance include projecting target language representations into the English space \citep{yang-etal-2022-enhancing,xu-etal-2023-language-representation}, and employing data augmentation \citep{zheng-etal-2021-consistency}. Notably, \citet{chi-etal-2021-infoxlm} and \citet{pan-etal-2021-multilingual} improve cross-lingual representation alignment in their InfoXLM PLM, employing contrastive learning and large-scale text corpora during pretraining. In contrast, our work focuses on data-efficient cross-lingual transfer during task-specific fine-tuning, eliminating the need for an additional cross-lingual alignment step on self-supervised data. 

\textbf{Few-shot Learning} and FS-XLT involve training models with a minimal number of labeled examples, challenging models to generalize from this limited data \citep{nie-etal-2023-cross}. Techniques explored for FS-XLT include fine-tuning \citep{lauscher-etal-2020-zero}, in-context learning \citep{winata-etal-2022-cross}, and prompting \citep{schick-schutze-2021-exploiting}. Prompting has shown notable performance improvements, mitigating hyperparameter sensitivity and performance variance issues prevalent in fine-tuning and in-context learning techniques \citep{zhao-etal-2021-closer,schmidt-etal-2022-dont,winata-etal-2022-cross}.
Enhancements to prompting techniques have been made through the use of multilingual templates \citep{qi-etal-2022-enhancing} and label words \citep{huang-etal-2022-zero, zhou-etal-2023-enhancing}, further improving their applicability in multilingual settings.
\citet{schmidt-etal-2023-free} note the difficulty in comparing studies due to varying few-shot settings. They suggest that averaging model checkpoints during the few-shot fine-tuning phase provides a robust baseline for FS-XLT. Our study contributes to this line of research by combining prompting techniques with contrastive learning to improve few-shot cross-lingual transfer performance. 

\section{Methodology}
\label{sec:methodology}

Our methodology is predicated on the hypothesis that the cross-lingual performance gap in downstream tasks for low-resource languages is largely attributable to their underdeveloped representation spaces within PLMs. Building on previous studies, we propose that aligning the representation spaces of low-resource languages with the more robust English representation space can significantly enhance task performance \citep{yang-etal-2022-enhancing, xu-etal-2023-language-representation}. This approach aims to transfer the discriminatory information embedded within the English representation space to these lower-resource languages.
Prior research, including the works of \citet{chi-etal-2021-infoxlm} and \citet{yang-etal-2022-enhancing}, supports the effectiveness of this alignment during language modeling. However, these methods often rely on large-scale parallel corpora for extensive pretraining, making them infeasible for low-resource scenarios. To address this, we introduce \colap, a method specifically designed for few-shot cross-lingual transfer in low-resource contexts. \colap avoids introducing additional model parameters through prompting, and does not require large-scale labeled or self-supervised corpora.
% We hypothesize that task-specific model representations are less complex compared to representations used for language modeling, and can be transferred more efficiently in terms of data and computation. 
Our approach includes a task-agnostic representation contrastive learning objective, \xrcl, which aligns multilingual vector representations of translated inputs. Unlike InfoXLM's contrastive learning objective, our method does not use a memory queue for negative pairs or mixup sampling, enhancing resource efficiency \citep{chi-etal-2021-infoxlm}. In addition, we introduce a classification task-specific contrastive learning objective, \xccl, designed to transfer discriminative, class-specific features from high-resource to lower-resource languages, alleviating the dependency on parallel training datasets. Importantly, the additional computational complexity introduced by our contrastive learning objectives is dependent only on the number of samples per batch. Both \colap strategies are model-agnostic, making them applicable to any PLM.

\subsection{Prompt-based Training}

In prompt-based training approaches, tasks are reformulated as language modeling problems, where the model predicts tokens that serve as reference labels \citep{schick-schutze-2021-exploiting}.
Specifically, a template $\mathcal{T}$ is applied to each input $x$, resulting in a prompted input $x_{\text{prompt}} = \mathcal{T}(x)$. The model then predicts task-specific label tokens based on the context provided by $x_{\text{prompt}}$. For models trained with a causal language modeling objective, we utilize the hidden state of the end-of-sequence token (\eos) to predict the labels. In contrast, for models trained with masked language modeling objectives, such as XLM-R, at least one \texttt{<mask>} token is included in the prompt , which is used to predict the label tokens. To facilitate this prediction, labels are mapped to tokens in the model's vocabulary, represented as $\mathcal{R} \mapsto \mathcal{V}_{\mathcal{M}}$. Given an input $x$, the probability of predicting label $y$ is denoted as:
\begin{equation*}
\begin{split}
    p(y \mid x)=p(\eos=w_{v} \mid x_{\text{prompt}}) \\
    =\frac{\exp{ \left( w_{v} \cdot h_{\eos} \right)}}{\sum_{v' \in \mathcal{V}}{\exp{ \left(w_{v'} \cdot h_{\eos} \right)}}},
\end{split}
\end{equation*}
where $h_{\eos}$ is the hidden vector of the $\eos$ token\footnote{In masked language models, label tokens are predicted at the positions of the \texttt{<mask>} tokens rather than at the \eos token.}, and $w_v$ denotes the pre-softmax vector corresponding to $v \in \mathcal{V}$ \citep{schick-schutze-2021-exploiting}. We employ English label words and language agnostic prompt templates, included in Table \ref{tab:templates} (appendix).

\subsection{Contrastive Learning Objectives}

\begin{figure}
% \vspace{-5pt}
    \centering
    \includegraphics[width=0.6\linewidth]{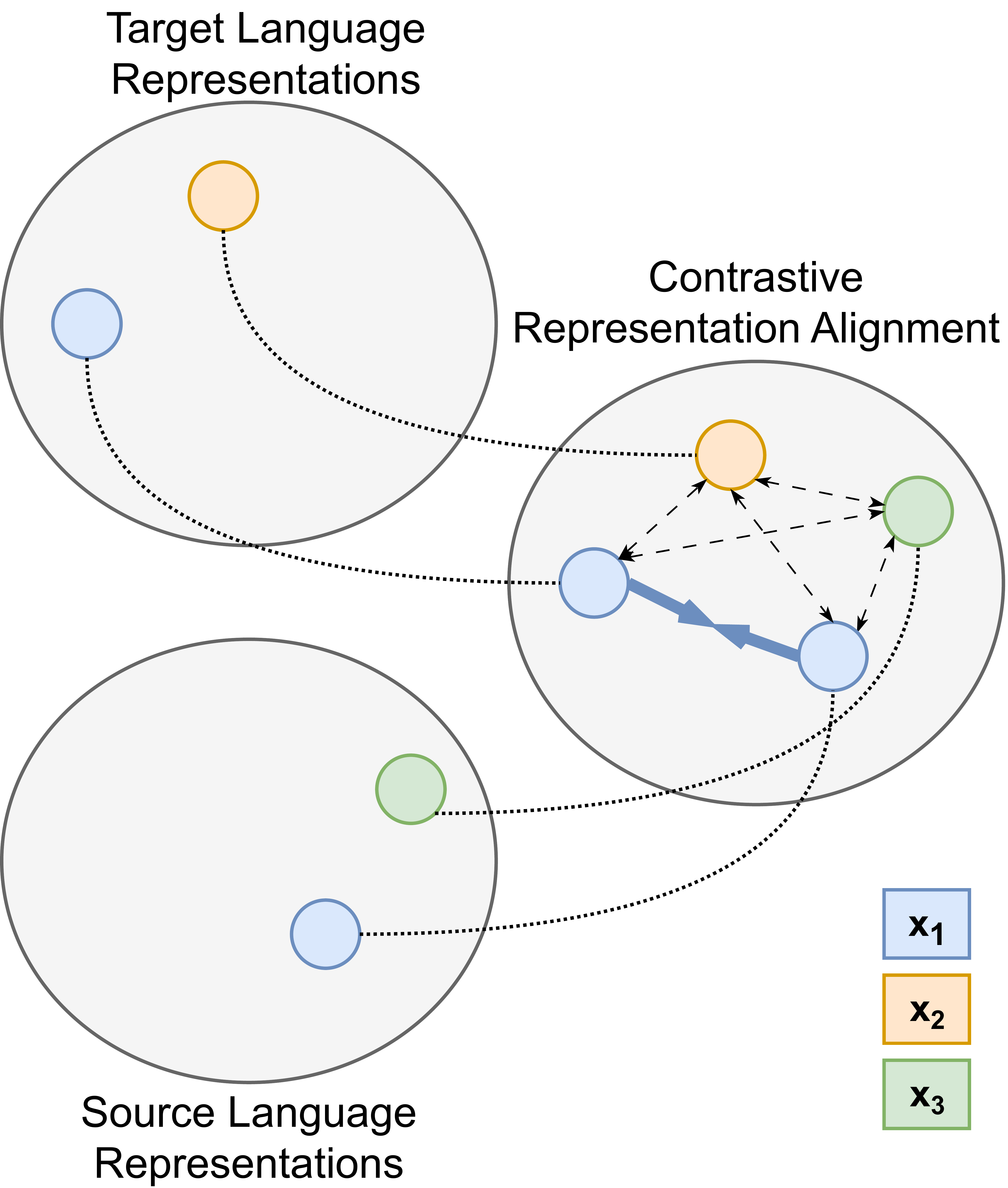}
    \caption{Illustration of cross-lingual contrastive representation alignment using the $\mathit{XRCL}$ objective.}
    \label{fig:xrcl}
    \vspace{-10pt}
\end{figure}

The \textbf{cross-lingual representation contrastive loss} term, abbreviated as $\mathcal{L}_{\xrclmath}$, aims to align the latent representations of instances in the target language with their counterparts in the source language. To achieve this, we utilize a model $\mathcal{M}$ to generate vector representations $r_i = \mathcal{M}(x_i)$ for each input sequence $x_i$ sampled along with its class label $y_i$ from the training set. We obtain these representations by prompting the PLM and extracting the hidden state of the \eos token.
We treat the target language training dataset ($D_T$) and the source language training dataset ($D_S$) as parallel corpora, meaning that each instance $x_{i,T}$ in $D_T$ has a direct translation $x_{i,S}$ in $D_S$.

For a given instance representation in the target language $r_{i,T}$, its direct translation in the source language $r_{i,S}$, forms a positive pair. Consequently, all other instances within the target language dataset are treated as negative pairs. This is formulated as follows:
\vspace{-5pt} \begin{equation*}
\begin{split}
    & r_i^+ = \{ r_j \, | \, j = i, j \in D_S \} \\
    & r_i^- = \{ r_j \, | \, j \neq i, j \in D_S \}
\end{split}
\end{equation*}

The \xrcl objective maximizes the similarity between these cross-lingual instance representations (positive pairs) while minimizing their similarity to other instances (negative pairs) \citep{oord2019representation}. We compute the cross-lingual representation contrastive loss, $\mathcal{L}_{\xrclmath}$, for each instance representation in the target language $r_{i,T}$ as follows:
\begin{equation*}
    \mathcal{L}_{\mathit{XRCL}} = \sum_{i=1}^{N}-log\frac{exp\left(\phi(r_{i,T}, r_i^+)/\tau \right)}{exp\left(\phi(r_{i,T},r_i^-)/\tau\right)},
    \label{eq:xrcl}
\end{equation*}
where $N$ is the number of instances in the target language's training set $D_T$. The parameter $\tau$ is a temperature scaling factor, and $\phi(r_i,r_i^-)$ calculates the cosine similarity between representation $r_i$ and each representation in $r_i^-$ using the formula $\sum_{j=1}{r_i \cdot r_j / |r_i| |r_j|}$.

The \textbf{cross-lingual class contrastive loss}, denoted as $\mathcal{L}_{\xcclmath}$, addresses the limitations of related cross-lingual transfer methods, which often assume that contextual information is accurately preserved through direct translations between the source and target languages. In practice, translating instances requires careful adaptation of culture-specific terminology and evaluation metrics, which incurs significant cost in the data collection process \citep{ponti-etal-2020-xcopa,freitag-etal-2022-natural,winata-etal-2023-nusax}. In contrast, our \xccl objective does not rely on parallel translations between the source and target languages. Instead, it aligns instances across different languages based on shared class labels. Specifically, for each representation $r_{i,T}$ from the target language training dataset, we construct positive pairs with representations of instances in the source language training dataset that have the same class label. Instances from the source language with different class labels are considered negative pairs. The formulation is as follows:
\begin{equation*}
\begin{split}
    & r_i^+ = \{ r_j \, | \, y_j = y_i, j \in D_S \} \\
    & r_i^- = \{ r_j \, | \, y_i \neq y_j, j \in D_S \}
\end{split}
\vspace{-5pt}
\end{equation*}

This approach aims to maximize the similarity between cross-lingual instance representations within the same class while minimizing similarity with different class representations. The key distinction of $\mathcal{L}_{\xcclmath}$ from $\mathcal{L}_{\xrclmath}$ lies in its assumption that class-specific information and can effectively be leveraged across languages. It posits that embeddings from instances within the same class should exhibit higher similarity compared to those from instances of different classes. The computation of the class-contrastive loss is analogous to \xrcl, as shown below:
\begin{equation*}
    \mathcal{L}_{\xcclmath} = \sum_{i=1}^{N}-log\frac{exp\left(\phi(r_{i,T}, r_i^+)/\tau \right)}{exp\left(\phi(r_{i,T},r_i^-)/\tau\right)}
    \label{eq:xccl}
\end{equation*}

\begin{figure}
% \vspace{-5pt}
    \centering
    \includegraphics[width=0.6\linewidth]{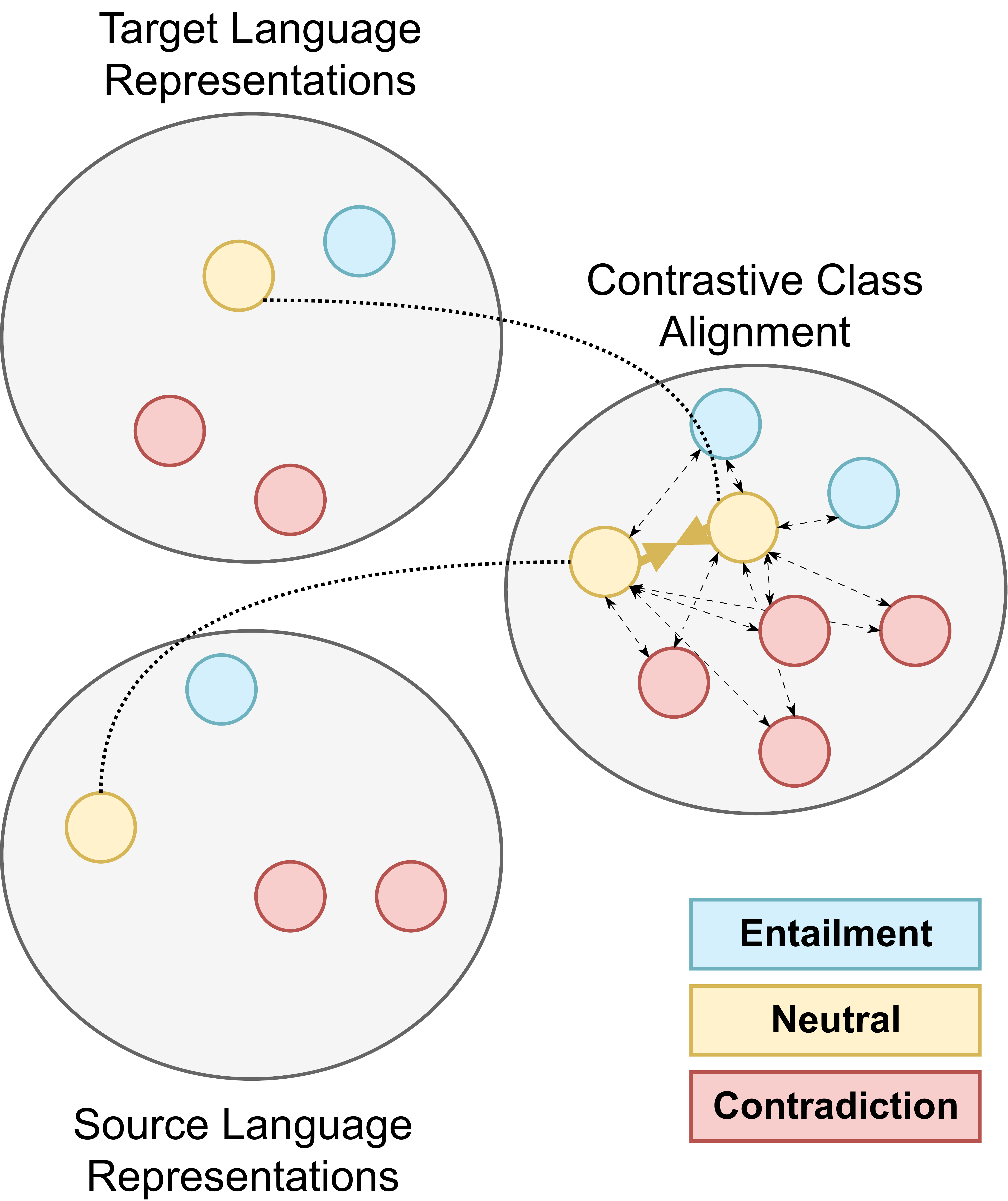}
    \caption{Illustration of cross-lingual class contrastive representation alignment using the \xccl objective.}
    \label{fig:xccl}
    \vspace{-10pt}
\end{figure}

\section{Experiments}
\label{sec:experiments}

\subsection{Models and Training}
\label{sec:model_training}

We investigate FS-XLT using models that are initially fine-tuned on downstream task data in a high-resource source language, such as English. During this phase, we employ the cross-entropy loss $\mathcal{L}_{CE}=-\log(z_y)$, where $z_y$ represents the post-softmax probability of the vocabulary token corresponding to label $y$. Subsequently, the model is adapted to the target language under FS-XLT.  During this few-shot adaptation phase, we integrate our contrastive loss terms. For the variant incorporating the cross-lingual representation contrastive objective \xrcl, the total loss is calculated as $\mathcal{L} = \mathcal{L}_{CE} + \mathcal{L}_{\xrclmath}$. Similarly, for the class contrastive objective variant \xccl, the total loss is $\mathcal{L} = \mathcal{L}_{\mathit{CE}} + \mathcal{L}_{\xcclmath}$. 

\sparagraphnodot{Training details}.
Unless specified otherwise, we adhere to training parameters that include a batch size of 64, a learning rate of 2e-5, and the AdamW optimizer \citep{loshchilov_2017_adamw}. Models are fine-tuned for 5 epochs on the downstream task in English. For the few-shot adaptation to the target language, we train the models for 10 epochs using data from both the source and target languages. Unlike some related studies, we do not use a dedicated validation set during few-shot fine-tuning. This approach aligns with the few-shot premise, avoiding the need for additional labeled data that a validation set would require \citep{schmidt-etal-2023-free}. Consequently, all models are trained for a fixed number of epochs on the few-shot data. We argue that this reflects real-world conditions more accurately, as performance is not skewed by the size or quality of a validation dataset. 

\sparagraphnodot{Models}.
We evaluate a range of encoder-only and decoder-only PLMs that have been exposed to multilingual text during pretraining, including XLM-R Base (270 million parameters) \citep{conneau-etal-2020-unsupervised}, Gemma 2 (2 billion parameters) \citep{gemmateam2024}, and Mistral v0.3 (7 billion parameters) \citep{jiang2023mistral7b}. XLM-R is trained using full-fine-tuning, while we utilize 4-bit quantization and low-rank adapters \citep{hu_2021_lora, dettmers2023qlora} with $r=16$ and $alpha=32$ for the larger Gemma 2 and Mistral models.\footnote{To ensure consistency and control for potential exposure to the downstream tasks included in our benchmarks, we do not evaluate instruction fine-tuned model variants.} 

\sparagraphnodot{Baselines}.
We include relevant benchmarks proposed in related studies. To ensure fair model comparisons, we integrate relevant benchmarks in our XLT settings, with all models being reimplemented for consistency.

\noindent Regular fine-tuning (\textbf{FT}) directly predicts class labels from sequence representations, in contrast to prompting approaches that use tokens as reference labels \citep{lauscher-etal-2020-zero,schmidt-etal-2022-dont}. The model is fine-tuned using the standard cross-entropy loss $\mathcal{L}_{\mathit{CE}}$.

\noindent Checkpoint Averaging (\textbf{CA}), introduced by \citet{schmidt-etal-2023-free}, is a strong baseline for FS-XLT. It builds upon regular fine-tuning by averaging model weights during few-shot adaptation.

\noindent In the prompt-learning from cross-lingual templates (\textbf{PCT}) framework \citet{qi-etal-2022-enhancing} augment model inputs with multilingual prompt templates, optimizing the model using both the $\mathcal{L}_{\mathit{CE}}$ objective for non-augmented inputs and the $\mathcal{L}_{\mathit{CE}'}$ objective for augmented inputs. Additionally, they introduce a consistency loss term based on the Kullback-Leibler divergence between predicted token probabilities. In contrast to \colap, PCT is applied during both supervised fine-tuning and few-shot adaptation phases, which requires the target languages to be known and reduces the modularity of the fine-tuned language model.

\noindent In-context Learning (\textbf{ICL}) incorporates few-shot examples directly into the input prompt, allowing the model to utilize these examples without updating its parameters. This approach significantly extends the prompt length during inference.\footnote{Due to context length and memory limitations, we benchmark ICL only for the $K=5$ few-shot setting and discuss the results in Section~\ref{sec:results}.}

\begin{table*}
\vspace{-12pt}
\centering
{
\def\arraystretch{0.99}
\fontsize{7.4pt}{7.3pt}\selectfont
    \begin{tabularx}{\textwidth}{lc!{\vrule}YYY!{\vrule}YYY!{\vrule}YYY}
    \toprule
    \multicolumn{1}{l}{\textbf{Model}} &
    \multicolumn{1}{c}{\textbf{K}} &
    \multicolumn{3}{c}{\textbf{XNLI}} &
    \multicolumn{3}{c}{\textbf{AmNLI}} &
    \multicolumn{3}{c}{\textbf{MultiTACRED}} \\
    \cmidrule(lr){3-5} \cmidrule(lr){6-8} \cmidrule(lr){9-11}
    & & \textsc{XLM-R} & \textsc{Gemma 2B} & \textsc{Mistral 7B} & \textsc{XLM-R} & \textsc{Gemma 2B} & \textsc{Mistral 7B} & \textsc{XLM-R} & \textsc{Gemma 2B} & \textsc{Mistral 7B} \\
    
    \midrule
    
    FT & \multirow{3}{*}{0} 
    & $73.21_{\pm1.0}$ & $69.32_{\pm0.9}$ & $61.98_{\pm1.3}$ 
    & $36.71_{\pm1.4}$ & $40.99_{\pm0.2}$ & $37.60_{\pm0.2}$ 
    & $24.57_{\pm1.3}$ & $14.55_{\pm0.5}$ & $10.96_{\pm0.5}$\\
    PCT & 
    & $72.97_{\pm1.2}$ & $71.81_{\pm0.7}$ & $67.68_{\pm1.4}$ 
    & $37.98_{\pm0.9}$ & $41.27_{\pm0.3}$ & $37.37_{\pm0.2}$ 
    & $45.87_{\pm2.3}$ & $34.16_{\pm0.8}$ & $34.05_{\pm1.6}$\\
    \rowcolor{lightblue} \colap & 
    & $73.15_{\pm1.2}$ & $\mathbf{74.35}_{\pm0.7}$ & $66.23_{\pm1.2}$ 
    & $35.79_{\pm0.7}$ & $\mathbf{42.73}_{\pm0.3}$ & $37.18_{\pm0.2}$ 
    & $\mathbf{51.58}_{\pm2.1}$ & $32.48_{\pm0.9}$ & $38.73_{\pm0.8}$\\

    \midrule

    FT & \multirow{5}{*}{5} 
    & $71.74_{\pm1.8}$ & $70.38_{\pm0.8}$ & $62.90_{\pm1.0}$ 
    & $39.57_{\pm1.4}$ & $41.41_{\pm0.2}$ & $37.58_{\pm0.2}$ 
    & $42.65_{\pm0.9}$ & $17.53_{\pm0.6}$ & $13.24_{\pm0.6}$\\
    CA & 
    & $73.21_{\pm1.2}$ & $70.38_{\pm0.8}$ & $62.90_{\pm1.0}$ 
    & $37.94_{\pm0.9}$ & $41.41_{\pm0.2}$ & $37.58_{\pm0.2}$ 
    & $38.94_{\pm1.5}$ & $17.53_{\pm0.6}$ & $13.24_{\pm0.6}$\\
    PCT & 
    & $72.52_{\pm1.3}$ & $72.45_{\pm0.7}$ & $67.94_{\pm1.2}$ 
    & $39.11_{\pm1.7}$ & $41.99_{\pm0.3}$ & $39.46_{\pm0.2}$ 
    & $69.01_{\pm1.2}$ & $44.00_{\pm0.9}$ & $43.35_{\pm1.6}$\\
    \rowcolor{lightblue} \colap w/ \xrcl & 
    & $73.56_{\pm1.2}$ & $\mathbf{74.59}_{\pm0.7}$ & $67.68_{\pm1.1}$ 
    & $40.01_{\pm1.4}$ & $\mathbf{42.88}_{\pm0.3}$ & $39.21_{\pm0.2}$ 
    & $\mathbf{69.26}_{\pm2.6}$ & $43.47_{\pm1.0}$ & $38.73_{\pm1.0}$\\
    \rowcolor{lightblue} \colap w/ \xccl & 
    & $73.04_{\pm1.2}$ & $74.54_{\pm0.7}$ & $67.58_{\pm0.0}$ 
    & $39.71_{\pm1.0}$ & $42.84_{\pm0.3}$ & $39.20_{\pm0.2}$ 
    & $69.18_{\pm2.6}$ & $42.91_{\pm1.0}$ & $37.96_{\pm1.0}$\\
    
    \midrule
    
    FT & \multirow{5}{*}{10} 
    & $72.47_{\pm1.4}$ & $70.53_{\pm0.8}$ & $63.02_{\pm1.1}$ 
    & $42.17_{\pm1.4}$ & $41.24_{\pm0.2}$ & $37.88_{\pm0.2}$
    & $43.91_{\pm1.1}$ & $18.35_{\pm0.6}$ & $13.83_{\pm0.6}$\\
    CA & 
    & $73.22_{\pm0.9}$ & $70.53_{\pm0.8}$ & $63.02_{\pm1.0}$ 
    & $40.28_{\pm0.5}$ & $41.24_{\pm0.2}$ & $37.88_{\pm0.2}$ 
    & $40.82_{\pm0.8}$ & $18.35_{\pm0.6}$ & $13.83_{\pm0.6}$\\
    PCT & 
    & $72.92_{\pm0.6}$ & $72.49_{\pm0.7}$ & $68.11_{\pm1.2}$ 
    & $41.26_{\pm0.9}$ & $42.41_{\pm0.3}$ & $40.01_{\pm0.2}$ 
    & $70.19_{\pm3.2}$ & $45.52_{\pm1.0}$ & $46.63_{\pm1.6}$\\
    \rowcolor{lightblue} \colap w/ \xrcl & 
    & $73.69_{\pm0.9}$ & $74.52_{\pm0.7}$ & $67.47_{\pm1.1}$ 
    & $40.69_{\pm1.0}$ & $42.36_{\pm0.3}$ & $39.29_{\pm0.2}$ 
    & $70.16_{\pm3.8}$ & $49.12_{\pm0.8}$ & $41.98_{\pm0.9}$\\
    \rowcolor{lightblue} \colap w/ \xccl & 
    & $73.15_{\pm0.9}$ & $\mathbf{74.55}_{\pm0.7}$ & $67.37_{\pm1.1}$ 
    & $41.09_{\pm1.2}$ & $\mathbf{42.67}_{\pm0.3}$ & $39.32_{\pm0.2}$ 
    & $\mathbf{70.32}_{\pm3.5}$ & $47.43_{\pm0.8}$ & $41.94_{\pm0.9}$\\
    
    \midrule

    FT & \multirow{5}{*}{50} 
    & $73.54_{\pm0.8}$ & $73.77_{\pm0.6}$ & $65.51_{\pm1.1}$ 
    & $\mathbf{44.76}_{\pm0.5}$ & $41.68_{\pm0.2}$ & $38.23_{\pm0.3}$ 
    & $46.88_{\pm1.5}$ & $29.45_{\pm0.9}$ & $20.68_{\pm0.8}$\\
    CA & 
    & $73.55_{\pm0.8}$ & $73.35_{\pm0.7}$ & $65.25_{\pm1.1}$ 
    & $42.31_{\pm0.5}$ & $41.66_{\pm0.2}$ & $38.08_{\pm0.3}$ 
    & $46.22_{\pm0.5}$ & $28.44_{\pm0.9}$ & $19.88_{\pm0.8}$\\
    PCT & 
    & $73.72_{\pm0.5}$ & $74.53_{\pm0.6}$ & $69.13_{\pm1.2}$ 
    & $43.38_{\pm0.6}$ & $43.64_{\pm0.3}$ & $41.94_{\pm0.2}$ 
    & $\mathbf{75.36}_{\pm1.3}$ & $62.14_{\pm0.8}$ & $63.33_{\pm1.3}$\\
    \rowcolor{lightblue} \colap w/ \xrcl & 
    & $73.88_{\pm0.7}$ & $\mathbf{75.81}_{\pm0.6}$ & $68.95_{\pm1.1}$ 
    & $41.98_{\pm0.4}$ & $43.14_{\pm0.3}$ & $39.84_{\pm0.2}$ 
    & $73.98_{\pm2.1}$ & $59.76_{\pm0.9}$ & $62.01_{\pm1.1}$\\
    \rowcolor{lightblue} \colap w/ \xccl & 
    & $73.81_{\pm0.5}$ & $75.73_{\pm0.6}$ & $68.78_{\pm1.1}$ 
    & $41.98_{\pm0.4}$ & $43.21_{\pm0.3}$ & $40.02_{\pm0.2}$ 
    & $73.98_{\pm2.1}$ & $57.90_{\pm0.9}$ & $61.31_{\pm1.1}$\\
    
    \midrule

    FT & \multirow{5}{*}{100} 
    & $73.87_{\pm0.6}$ & $74.38_{\pm0.6}$ & $67.07_{\pm1.1}$ 
    & $\mathbf{45.94}_{\pm0.4}$ & $42.76_{\pm0.2}$ & $38.57_{\pm0.3}$ 
    & $48.87_{\pm1.7}$ & $45.97_{\pm1.4}$ & $42.11_{\pm1.9}$\\
    CA & 
    & $73.95_{\pm0.5}$ & $74.40_{\pm0.6}$ & $66.91_{\pm1.1}$ 
    & $45.39_{\pm0.4}$ & $42.70_{\pm0.2}$ & $38.50_{\pm0.3}$ 
    & $48.41_{\pm1.6}$ & $42.82_{\pm1.2}$ & $41.06_{\pm1.8}$\\
    PCT & 
    & $73.80_{\pm0.5}$ & $75.25_{\pm0.6}$ & $69.57_{\pm1.1}$ 
    & $45.23_{\pm0.4}$ & $43.21_{\pm0.3}$ & $42.69_{\pm0.2}$ 
    & $\mathbf{78.26}_{\pm8.7}$ & $71.91_{\pm0.7}$ & $74.67_{\pm1.1}$\\
    \rowcolor{lightblue} \colap w/ \xrcl & 
    & $74.09_{\pm0.3}$ & $75.89_{\pm0.6}$ & $69.62_{\pm1.0}$ 
    & $44.79_{\pm0.3}$ & $43.64_{\pm0.3}$ & $40.27_{\pm0.3}$ 
    & $77.29_{\pm3.5}$ & $74.82_{\pm0.8}$ & $73.97_{\pm1.1}$\\
    \rowcolor{lightblue} \colap w/ \xccl & 
    & $74.03_{\pm0.4}$ & $\mathbf{75.94}_{\pm0.6}$ & $69.43_{\pm1.1}$ 
    & $45.18_{\pm0.2}$ & $43.59_{\pm0.3}$ & $40.25_{\pm0.3}$ 
    & $77.06_{\pm3.6}$ & $71.29_{\pm0.9}$ & $74.18_{\pm1.1}$\\
    
    \midrule

    FT & \multirow{5}{*}{250} 
    & $73.35_{\pm0.7}$ & $74.51_{\pm0.6}$ & $67.58_{\pm1.1}$ 
    & $48.53_{\pm1.2}$ & $44.07_{\pm0.3}$ & $41.43_{\pm0.3}$ 
    & $55.12_{\pm5.1}$ & $82.47_{\pm1.2}$ & $89.75_{\pm1.0}$\\
    CA & 
    & $73.63_{\pm0.6}$ & $74.58_{\pm0.6}$ & $67.45_{\pm1.1}$ 
    & $48.20_{\pm1.1}$ & $43.80_{\pm0.3}$ & $40.54_{\pm0.3}$ 
    & $53.69_{\pm4.3}$ & $84.81_{\pm1.2}$ & $88.64_{\pm1.0}$\\
    PCT & 
    & $73.03_{\pm0.6}$ & $75.74_{\pm0.6}$ & $70.25_{\pm1.1}$ 
    & $47.73_{\pm1.1}$ & $45.28_{\pm0.3}$ & $45.21_{\pm0.3}$ 
    & $82.19_{\pm4.5}$ & $81.60_{\pm0.6}$ & $87.98_{\pm0.5}$\\
    \rowcolor{lightblue} \colap w/ \xrcl & 
    & $74.00_{\pm0.3}$ & $75.68_{\pm0.5}$ & $70.37_{\pm1.1}$ 
    & $\mathbf{48.64}_{\pm1.2}$ & $44.97_{\pm0.3}$ & $42.45_{\pm0.2}$ 
    & $82.16_{\pm7.7}$ & $87.48_{\pm0.7}$ & $\mathbf{90.68}_{\pm1.0}$\\
    \rowcolor{lightblue} \colap w/ \xccl & 
    & $73.60_{\pm0.6}$ & $\mathbf{75.96}_{\pm0.6}$ & $70.51_{\pm1.1}$ 
    & $48.44_{\pm0.7}$ & $44.90_{\pm0.3}$ & $42.56_{\pm0.2}$ 
    & $81.99_{\pm7.5}$ & $86.25_{\pm0.7}$ & $90.39_{\pm1.0}$\\
    
    \bottomrule
    \end{tabularx}
    }
    \caption{Average accuracy across all non-English languages on the XNLI, AmNLI, and MultiTACRED datasets. Fine-tuning (FT), checkpoint averaging (CA), prompt-learning from cross-lingual templates (PCT), and \colap models are evaluated in few-shot adaptation to the target language following fine-tuning on English task data. The best results per dataset and few-shot setting are highlighted in \textbf{bold}.}
    \label{tab:results}
    \vspace{-10pt}
\end{table*}

\subsection{Episode Sampling}
\label{sec:episode_sampling}

In the few-shot task adaptation, episodes are randomly sampled from the training dataset of the target language. Each episode consists of $K$ input sequences along with corresponding labels. The value of $K$ is varied among the set $\{5, 10, 50, 100, 250\}$. Episodes comprise $K$ uniformly sampled instances from the available class labels $N_Y$. In scenarios where $K \mod N_Y \neq 0$ the remaining instances are randomly selected.

For the computation of our cross-lingual contrastive loss terms, an episode is constructed to include $K$ instances from the training dataset in the target language $D_T$ and an additional $K$ instances from the training dataset in the source language $D_S$. It is important to note that we assume $D_T$ and $D_S$ to be parallel corpora, meaning that all instances present in the training dataset of the target language $D_T$ are also available in the training dataset of the source language $D_S$.

\subsection{Datasets and Languages}

\sparagraphnodot{XNLI} contains 15 languages, each containing 7,500 examples. In natural language inference, the goal is to determine if a given hypothesis entails, contradicts, or is neutral relative to a premise. Initially annotated in English, the dataset was expanded to other languages using machine translation \citep{conneau-etal-2018-xnli}. We assess model performance using the accuracy metric.

\sparagraphnodot{AmericasNLI (AmNLI)} extends the XNLI dataset by adding 10 indigenous languages from the Americas, categorized as very low-resource languages due to their minimal presence in large-scale text corpora. The dataset is human-translated based on the Spanish XNLI. It includes 750 development and 750 test instances per language, which we utilize as our training and testing sets, respectively \citep{ebrahimi-etal-2022-americasnli}. Performance evaluation on AmericasNLI is based on accuracy.\footnote{Otomí ``oto'' has less than 250 samples and therefore is not included in the $K=250$ setting.}

\sparagraphnodot{MultiTACRED} is a multilingual relation extraction dataset covering 12 languages, created via machine translation of English annotations \citep{hennig-etal-2023-multitacred}. The primary task in sentence-level relation extraction involves categorizing a given set of entities into one of 41 distinct relation types. We utilize instances from the development set to construct few-shot episodes and assess model performance based on accuracy.

\section{Results and Discussion}
\label{sec:results}

We summarize our results in Table~\ref{tab:results}. For each language in the datasets, we averaged accuracy metrics across five random seeds.

Our findings confirm our initial hypothesis that discriminative task-specific information can be effectively transferred from English to lower-resource languages using only a few labeled examples. Our \colap method consistently improves downstream performance over zero-shot baselines on natural language inference and relation extraction tasks across all evaluated few-shot settings. This improvement is observed for languages included during the pretraining of the PLMs (such as those in XNLI and MultiTACRED) as well as for unseen languages, like those in AmNLI. 
Furthermore, the performance gains provided by \colap are independent of the PLM architecture. Both encoder-only models (e.g., XLM-R) and decoder-only models (e.g., Gemma and Mistral) show increased FS-XLT performance when utilizing \colap. In the lowest-resource setting of $K=5$, \colap demonstrates strong performance, outperforming all baseline approaches for the XLM-R and Gemma PLMs and performing on par with PCT for the Mistral PLM. These performance benefits remain consistent across different few-shot settings. Even in the most resource-rich scenario of $K=250$ exemplars, \colap surpasses the performance of its benchmarks, except in the AmNLI dataset for Gemma and Mistral. Even in the high-resource setting of $K=250$ exemplars, \colap exceeds the performance of its benchmarks. When compared to the strongest baseline, PCT (which also uses prompting), \colap achieves average performance gains of up to 1.84\% for Gemma 2.

Our results demonstrate that prompting approaches, including \colap and PCT, outperform traditional fine-tuning methods in the FS-XLT setting. This indicates that regular fine-tuning requires significantly more exemplars to achieve performance comparable to the prompting models on average. While \colap is the most data-efficient approach, all evaluated few-shot models provide substantial performance improvements over zero-shot settings, even with as few as $K=50$ exemplars. For the MultiTACRED dataset, with its large number of relation types, prompting-based approaches like \colap or PCT prove to be more suitable.

\begin{figure}[!tbp]
\vspace{-8pt}
  \centering
  \subfloat{\includegraphics[width=0.48\linewidth]{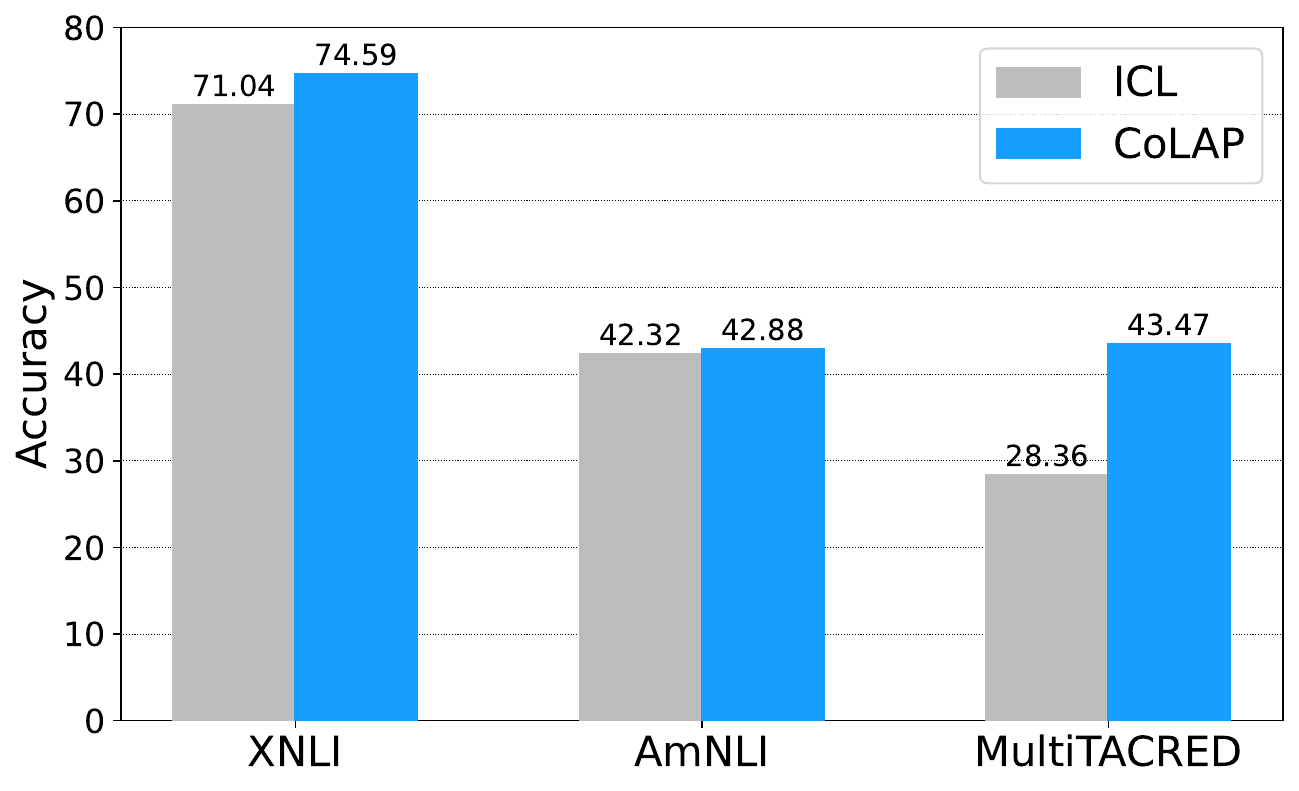}}
  \hfill
  \subfloat{\includegraphics[width=0.48\linewidth]{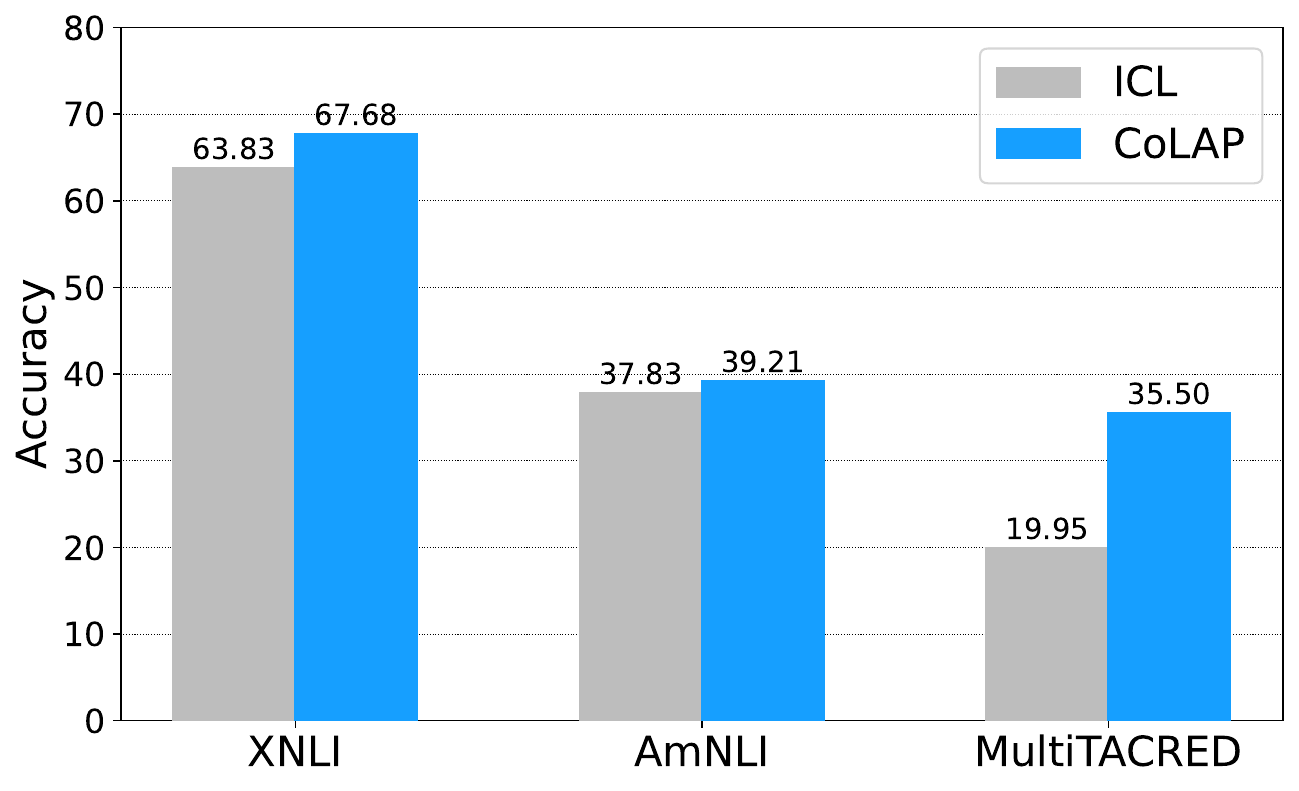}}
  \caption{Performance comparison between in-context learning (ICL) and our \colap \xrcl method on XNLI, AmNLI, and MultiTACRED datasets with $K=5$ exemplars for Gemma 2 (left) and Mistral (right).}
  \label{fig:results_icl}
  \vspace{-8pt}
\end{figure}

% Introduce ICL performance
In Figure \ref{fig:results_icl}, we compare the performance of in-context learning with our \colap method, focusing on decoder-only PLMs, specifically Gemma 2 and Mistral, in the $K=5$ few-shot setting. Both approaches start from the same prompting model fine-tuned on English task data. Our \colap method consistently outperforms ICL across all natural language understanding benchmarks, achieving average performance improvements of 6.41\% for Gemma 2 and 6.93\% for Mistral. The performance gap is particularly pronounced on the MultiTACRED dataset.
Additionally, we include ablation study results in Table \ref{tab:ablation} (appendix), where we evaluate the individual components of \colap and explore combinations with related approaches.

\begin{table*}
\vspace{-10pt}
\centering
{
\def\arraystretch{0.99}
\fontsize{8pt}{7.9pt}\selectfont
    \begin{tabularx}{\linewidth}{llc!{\vrule}*{2}{YYY!{\vrule}}YYY}
    \toprule
    \textbf{PLM} &
    \textbf{Model} &
    \multicolumn{1}{c}{\textbf{K}} &
    \multicolumn{3}{c}{\textbf{XNLI}} &
    \multicolumn{3}{c}{\textbf{AmNLI}} &
    \multicolumn{3}{c}{\textbf{MultiTACRED}} \\
    \cmidrule(lr){4-6} \cmidrule(lr){7-9} \cmidrule(lr){10-12}
    & & & Random & High & Low & Random & High & Low & Random & High & Low \\ 
    \midrule

    XLM-R & \colap w/ \xrcl & \multirow{2}{*}{5} & $ 73.56 $ & $ 74.41 $ & $ 74.22 $ & $ 40.01 $ & $ 38.86 $ & $ 38.81 $ & $ 69.26 $ & $ 72.22 $ & $ 69.62 $ \\
    & \colap w/ \xccl &  & $ 73.04 $ & $74.38 $& $74.12$ & $39.71$ & $39.32$ & $39.33$ & $69.18$ & $71.72$ & $69.36$ \\

    \midrule
    
    XLM-R & \colap w/ \xrcl & \multirow{2}{*}{10} & $73.69$ & $73.87$ & $74.08$ & $40.69$ & $39.52$ & $40.01$ & $70.16$ & $74.73$ & $70.87$ \\
    & \colap w/ \xccl &  & $73.15$ & $74.10$ & $73.85$ & $41.09$ & $39.40$ & $40.33$ & $70.32$ & $74.74$ & $69.82$ \\

    \bottomrule
    \end{tabularx}}
    \caption{Evaluation of exemplar selection using representation similarity scores, comparing ``High'' and ``Low'' similarity exemplars against random selection, with accuracy as the evaluation metric.}
    \label{tab:ablation_exemplars}
    \vspace{-10pt}
\end{table*}

\subsection{Enhancing FS-XLT Without Parallel Translations}

The results in Table \ref{tab:results} demonstrate that the \colap variants utilizing the \xccl objective effectively transfer class-specific features from English to low-resource languages. Notably, the \xccl variant enhances FS-XLT performance without relying on parallel translations of the few-shot examples, thereby simplifying the data annotation process. By eliminating the need for parallel data, \xccl avoids the challenges associated with translating difficult or culturally specific examples and allows for the selection of more natural few-shot instances that share the same class labels. Despite not using parallel translations, the \xccl variant exhibits only a minimal reduction of less than one percent in average few-shot performance ($K > 0$) compared to the \xrcl variant across all datasets and settings. While \xrcl outperforms \xccl in very low-resource settings, as few as $K=10$ exemplars are sufficient for \xccl to match or exceed the performance of the \xrcl variant.

\subsection{Layer Selection for Contrastive Learning}

\begin{figure}[!tbp]
\vspace{-8pt}
  \centering
  \subfloat{\includegraphics[width=0.48\linewidth]{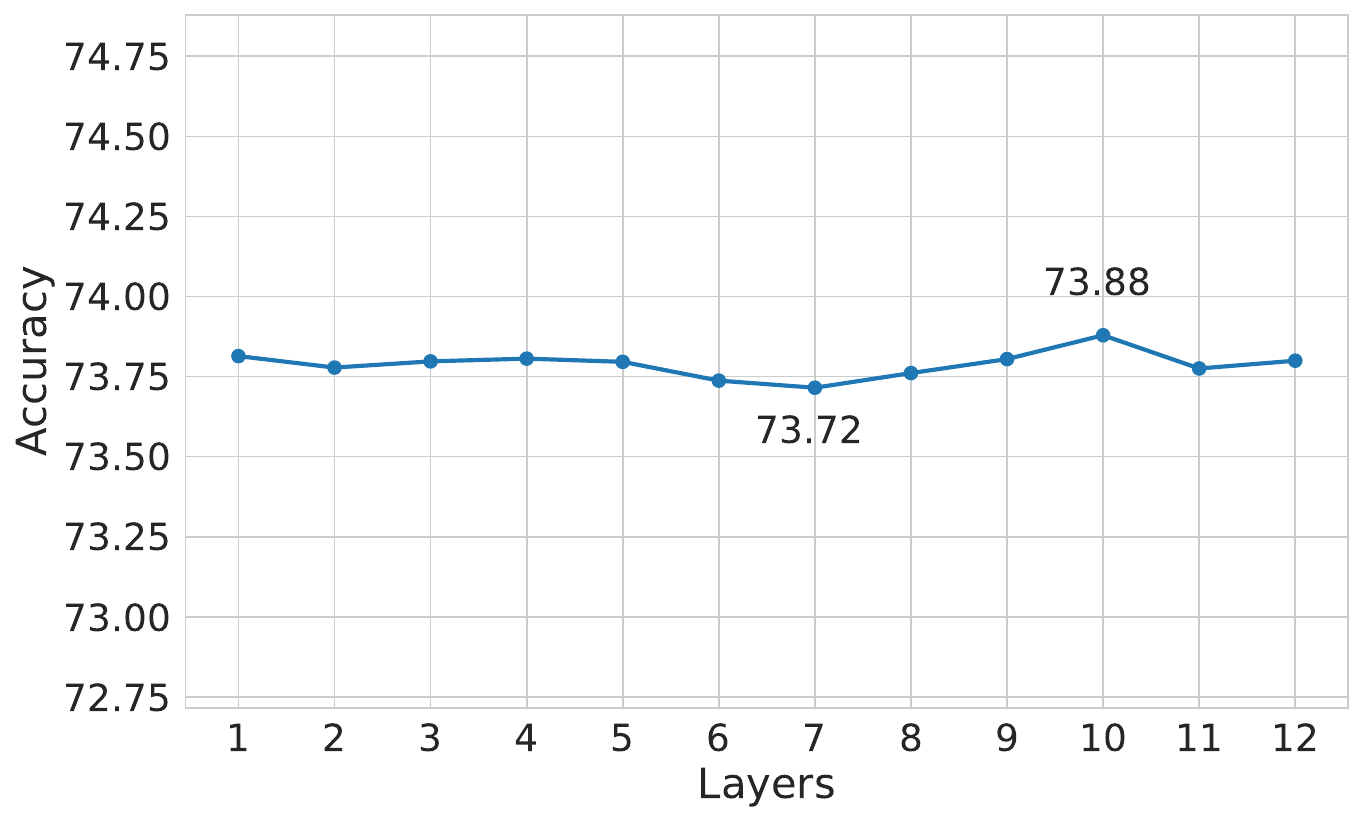}}
  \hfill
  \subfloat{\includegraphics[width=0.48\linewidth]{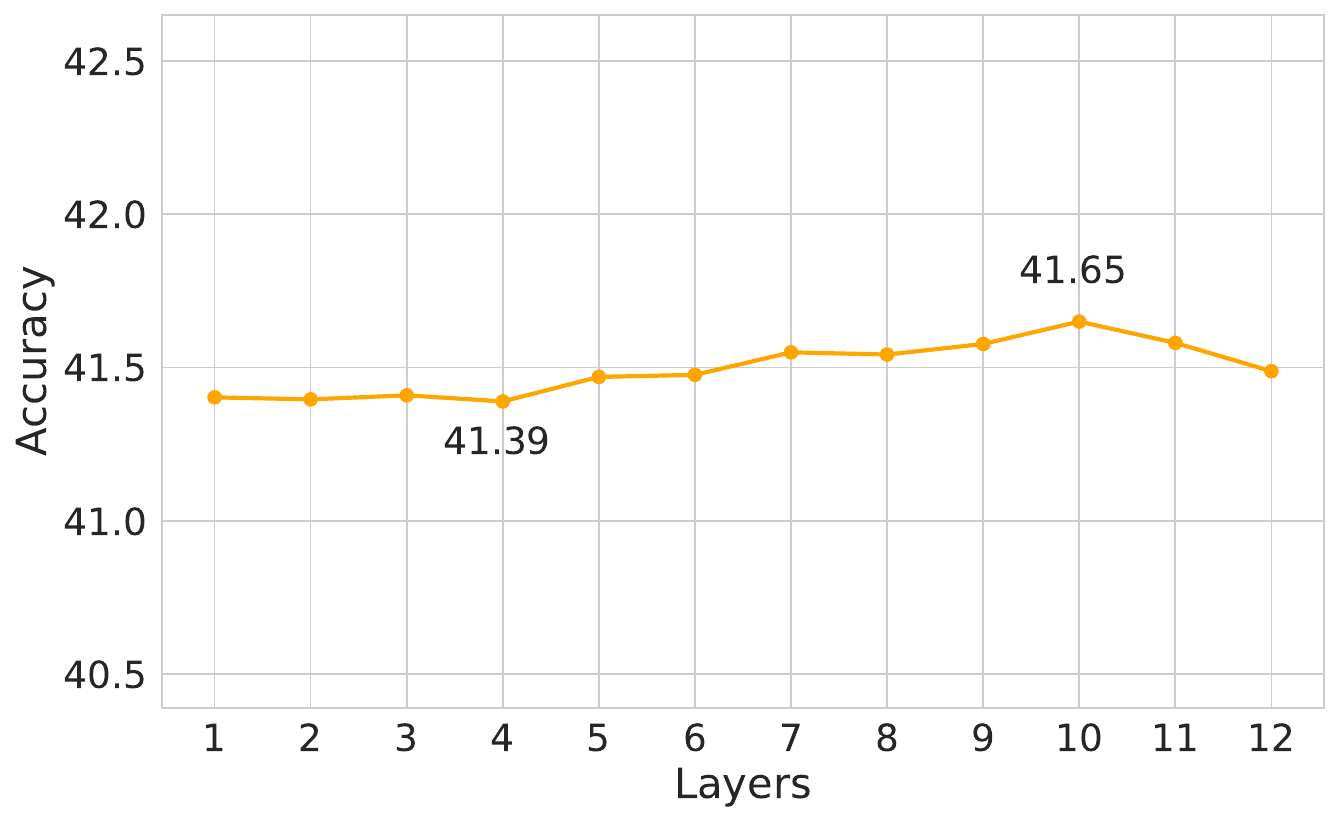}}
  \caption{Comparative performance of utilizing different layer representations for contrastive learning in the \colap \xrcl model with K=50 exemplars. Results are shown for XNLI (left) and AmNLI (right) datasets.}
  \label{fig:ablation_layers}
  \vspace{-8pt}
\end{figure}

To determine the most effective layer for contrastive representation alignment, we conduct experiments with the \colap \xrcl model, selecting $K=50$ exemplars. The findings, illustrated in Figure \ref{fig:ablation_layers}, reveal that applying contrastive representation learning at the 10th layer of XLM-R yields the best performance for NLI tasks\footnote{We note that these results may vary depending on the specific downstream tasks.}. This observation suggests that contrastive learning objectives benefit from the general-domain information captured in the middle layers of the model. In contrast, initial layers process low-level syntactic features, whereas the final layers are more focused on extracting discriminative features relevant for the classification task at hand. This finding diverges from that of \citet{chi-etal-2021-infoxlm}, who identified the 8th layer as optimal for the contrastive learning objectives used during the pretraining of InfoXLM.

\subsection{Few-Shot Exemplar Selection}

Building upon the performance of our class contrastive learning objective, we investigate the potential of utilizing class representation similarity for selecting few-shot exemplars. We hypothesize that transferring information from English to other languages via contrastive learning is more sample efficient when selecting exemplars with high intra-class similarity and low inter-class similarity.
Employing the XLM-R prompting model, fine-tuned on English task data, we extracted \eos representations from English training instances. We then created class prototypes by averaging these embeddings for each class.
For each instance $x_i$, we compute the cosine similarity to its own class prototype $P^+$ and dissimilarity to other class prototypes $P^-$, using the formula $s_i=\phi(x_i,P^+) + N_Y - \phi(x_i,P^-)$, where $N_Y$ is the total number of classes. The computational complexity of this procedure is determined by the pairwise similarity calculations, which depend on the number of instances and classes. We selected exemplars based on the highest or lowest similarity scores, focusing on sets of $K=5$ and $K=10$. Few-shot episodes consist of these selected exemplars in the source language and their translations in the target language.
Table \ref{tab:ablation_exemplars} shows that selecting exemplars based on representation similarity enhances efficiency for languages in XNLI and MultiTACRED, improving data efficiency by at least 50\%. Notably, with just $K=5$ exemplars selected based on our similarity criteria, \colap exceeds the performance of randomly selected sets of $K=250$ exemplars for XNLI.
However, for languages not included in the pretraining data, selecting exemplars based on similarity in English yields similar or worse results compared to random selection.

\section{Conclusion}
\label{sec:conclusion}

Our study showcases the effectiveness of contrastive learning in enhancing few-shot adaptation for PLMs. By aligning cross-lingual instance representations, we substantially improve performance in few-shot cross-lingual transfer on natural language inference and relation extraction tasks. In contrast to existing contrastive learning approaches, \colap does not require extensive pretraining on parallel multilingual corpora. Despite this, it achieves performance gains even with the smallest set of few-shot exemplars. We show that strong few-shot cross-lingual transfer can be accomplished without relying on parallel translations, which simplifies the data collection process and may reduce costs. Additionally, we demonstrate that selecting few-shot exemplars based on representation similarity enhances data efficiency for languages included in the PLM pretraining corpora.

\section{Limitations}
\label{sec:limitations}

In \colap, we introduce an effective strategy to narrow the cross-lingual transfer performance gap in few-shot cross-lingual transfer, tackling the challenge of imbalanced language representation in the pretraining data of current PLMs. Although our approach enhances data efficiency in improving downstream tasks, it does not directly resolve the underlying disparities that contribute to the cross-lingual transfer gap in PLMs.

A key consideration in \colap's implementation is the need for translations of the few-shot examples, predicated on the assumption that acquiring (machine) translations, especially in high-resource languages like English, is cost-effective and straightforward. Nonetheless, this requirement introduces an additional step in the process. While the class contrastive learning objective (\xccl) in \colap diminishes the dependency on parallel corpora, its counterpart, the cross-lingual representation contrastive learning objective (\xrcl), requires translated instance pairs. Furthermore, the advantage of employing \colap with the \xccl objective for not requiring parallel corpora during few-shot training, is limited to classification tasks.

\section*{Acknowledgements}
The resources and services used in this work were provided by the VSC (Flemish Supercomputer Center), funded by the Research Foundation - Flanders (FWO) and the Flemish Government.

\bibliography{anthology,custom}

\newpage
\appendix

\section{Training Details}
\label{sec:appendix_training}

Training is conducted on NVIDIA A6000, H100 and A100 GPUs. The process of fine-tuning XLM-R with prompt-based techniques on the English XNLI dataset takes approximately 10 hours on A6000 GPUs. For few-shot adaptation employing \colap across 14 languages from the XNLI dataset (English excluded) in the $K=250$ setting, the duration is about 30 minutes.

\begin{table*}
\centering
{
\def\arraystretch{0.99}
\fontsize{9pt}{8.9pt}\selectfont
    \begin{tabularx}{\linewidth}{lll}
    \toprule
    \textbf{Task} &
    \textbf{PLM} &
    \textbf{Template} \\
    \midrule

    \multirow{2}{*}{XNLI, AmNLI} & XLM-R & \texttt{<premise>} \texttt{<mask>}, \texttt{<hypothesis>} \\
    & Gemma, Mistral & \texttt{<premise>} \texttt{<hypothesis>}, \eos \\

    \midrule
    
    \multirow{2}{*}{MultiTACRED} & XLM-R & \texttt{<sentence>} \texttt{<E1>} \texttt{<mask>} \texttt{<E2>} \\
    & Gemma, Mistral & \texttt{<sentence>} \texttt{<E1>} \texttt{<E2>}, \eos  \\

    \bottomrule
    \end{tabularx}}
    \caption{Language agnostic prompt templates for encoder-only and decoder-only language models on natural language inference and relation extraction tasks.}
    \label{tab:templates}
\end{table*}

In Table \ref{tab:templates}, we display the language agnostic prompting templates used in our \colap approach. For NLI tasks, we utilize the label words ``Yes'', ``Maybe'', and ``No'' are used, in line with \citet{schick-schutze-2021-exploiting}. For the PCT approach, we utilize the multilingual prompt templates introduced by \citet{qi-etal-2022-enhancing}, such as ``\texttt{<premise>} Question:\texttt{<hypothesis>}? Answer:\eos''. 

For relation extraction tasks, we replace the entity markers \texttt{<E1>} and \texttt{<E2>} with the respective entities mentioned in the context of the sentence. We adapt the multilingual PCT template to the relation extraction task ``\texttt{<sentence>} Relation:\texttt{<E1>}, \texttt{<E2>}? Answer:\eos'' and machine translate the English template using Google translate \citep{qi-etal-2022-enhancing}. 

For MultiTACRED, we augment the input sentences for benchmarked models by including entity marker tokens \texttt{<E1>}, \texttt{</E1>}, \texttt{<E2>}, and \texttt{</E2>} indicating the position of the entities in the context. We reduce the number relation types from 41 to 31 by merging overlapping class labels for all models evaluated on MultiTACRED. The corresponding label mapping and template files are available at \url{https://github.com/pnborchert/CoLAP}.
% https://anonymous.4open.science/r/CoLAP-B2BE

To accurately reflect the real-world few-shot performance, we consistently train all models for 10 epochs on the few-shot episodes. We observe that \colap models trained on episodes with $K$ greater than 100 instances gain performance when trained up to 50 epochs.

\section{Ablation Studies}

Table \ref{tab:ablation} displays the performance results of integrating various strategies from related work with our \colap method. We evaluate these combined model variants on both small ($K=5$) and large ($K=250$) few-shot settings. 
Our analysis confirms that the contrastive learning objectives are critical to \colap's performance, as removing them leads to a noticeable drop in performance. Additionally, combining both the \xrcl and \xccl objectives yields better performance compared to using only one, though this requires parallel translations of the few-shot instances.
The results also show that checkpoint averaging (CA) enhances robustness, particularly for larger PLMs, improving the stability of \colap models. While combining multilingual prompt templates through PCT with \colap shows potential, it underperforms compared to other approaches, suggesting that further research is needed to combine these methods.

\begin{table*}
\centering
{
\def\arraystretch{0.99}
\fontsize{7.4pt}{7.3pt}\selectfont
    \begin{tabularx}{\textwidth}{lc!{\vrule}YYY!{\vrule}YYY!{\vrule}YYY}
    \toprule
    \multicolumn{1}{l}{\textbf{Model}} &
    \multicolumn{1}{c}{\textbf{K}} &
    \multicolumn{3}{c}{\textbf{XNLI}} &
    \multicolumn{3}{c}{\textbf{AmNLI}} &
    \multicolumn{3}{c}{\textbf{MultiTACRED}} \\
    \cmidrule(lr){3-5} \cmidrule(lr){6-8} \cmidrule(lr){9-11}
    & & \textsc{XLM-R} & \textsc{Gemma 2B} & \textsc{Mistral 7B} & \textsc{XLM-R} & \textsc{Gemma 2B} & \textsc{Mistral 7B} & \textsc{XLM-R} & \textsc{Gemma 2B} & \textsc{Mistral 7B} \\
    
    \midrule
    
    \colap w/ \xrcl & \multirow{5}{*}{5} 
    & $ 73.56 $ & $ 74.59 $ & $ 67.68 $
    & $ 40.01 $ & $ 42.88 $ & $ 39.21 $ 
    & $ 69.26 $ & $ 43.47 $ & $ 35.50 $\\
    w/o \xrcl & 
    & $ 72.89 $ & $ 72.48 $ & $ 67.34 $
    & $ 39.87 $ & $ 40.33 $ & $ 38.83 $ 
    & $ 68.90 $ & $ 41.05 $ & $ 35.09 $\\
    w/ \xccl & 
    & $ 72.60 $ & $ 74.42 $ & $ 67.59 $
    & $ 39.78 $ & $ 42.70 $ & $ \underline{39.29} $ 
    & $ 69.28 $ & $ 39.79 $ & $ 35.39 $\\
    w/ CA & 
    & $ \underline{73.75} $ & $ 74.58 $ & $ 67.65 $
    & $ 39.03 $ & $ 42.60 $ & $ 39.14 $ 
    & $ 64.79 $ & $ 43.41 $ & $ \underline{38.23} $\\
    w/ PCT & 
    & $ 72.95 $ & $ 74.22 $ & $ 67.32 $
    & $ 38.16 $ & $ 42.05 $ & $ 38.91 $ 
    & $ \underline{69.87} $ & $ 43.18 $ & $ 30.67 $\\

    \midrule

    \colap w/ \xrcl & \multirow{5}{*}{250} 
    & $ 74.00 $ & $ 75.68 $ & $ 70.37 $
    & $ 48.64 $ & $ 44.97 $ & $ 42.45 $ 
    & $ 82.16 $ & $ 87.48 $ & $ 90.80 $\\
    w/o \xrcl & 
    & $ 73.72 $ & $ 73.83 $ & $ 70.18 $
    & $ 41.74 $ & $ 43.68 $ & $ 41.95 $ 
    & $ 81.95 $ & $ 83.99 $ & $ 89.97 $\\
    w/ \xccl & 
    & $ 73.78 $ & $ \underline{76.03} $ & $ \underline{70.59} $
    & $ 48.28 $ & $ 44.40 $ & $ \underline{42.54} $ 
    & $ 82.02 $ & $ \underline{87.90} $ & $ 90.77 $\\
    w/ CA & 
    & $ 73.89 $ & $ \underline{75.79} $ & $ \underline{70.75} $
    & $ 48.11 $ & $ 44.84 $ & $ 42.31 $ 
    & $ 81.62 $ & $ 87.26 $ & $ 87.35 $\\
    w/ PCT & 
    & $ 73.45 $ & $ 75.38 $ & $ 70.20 $
    & $ 47.68 $ & $ 44.57 $ & $ 42.13 $ 
    & $ \underline{84.97} $ & $ 86.38 $ & $ 87.15 $\\

    \bottomrule
    \end{tabularx}}
    \caption{Model variants with (w/) or without (w/o) indicated architectural changes. Results that improve performance over the \colap variant with \xrcl are \underline{underlined}.}
    \label{tab:ablation}
\end{table*}

\section{Detailed Results}

The aggregated results presented in Table \ref{tab:results} are broken down by PLMs and individual languages in Tables \ref{tab:results_xlmr_xnli}, \ref{tab:results_gemma_xnli}, \ref{tab:results_mistral_xnli}, \ref{tab:results_xlmr_anli}, \ref{tab:results_gemma_anli}, \ref{tab:results_mistral_anli},\ref{tab:results_xlmr_tacred}, \ref{tab:results_gemma_tacred}, and \ref{tab:results_mistral_tacred} for detailed analysis. Given that English serves as the source language in our study, its accuracy scores are excluded from the average performance calculations.

\begin{table*}
\centering
{
\def\arraystretch{0.99}
\fontsize{5.5pt}{5.4pt}\selectfont
\begin{tabularx}{\linewidth}{lc!{\vrule}>{\columncolor{gray!25}}Y*{14}{Y}!{\vrule}Y}
    \toprule
    \textbf{Model} &
    \textbf{K} &
    \textbf{en} &
    \textbf{ar} &
    \textbf{bg} &
    \textbf{de} &
    \textbf{el} &
    \textbf{es} &
    \textbf{fr} &
    \textbf{hi} &
    \textbf{ru} &
    \textbf{sw} &
    \textbf{th} &
    \textbf{tr} &
    \textbf{ur} &
    \textbf{vi} &
    \textbf{zh} &
    \textbf{Avg.} \\

    \midrule

    FT  &  \multirow{3}{*}{0}  & $84.07$ & $ 71.38 $ & $ 77.25 $ & $ 76.05 $ & $ 75.42 $ & $ 78.66 $ & $ 77.23 $ & $ 69.22 $ & $ 75.69 $ & $ 65.10 $ & $ 72.33 $ & $ 72.62 $ & $ 65.66 $ & $ 74.75 $ & $ 73.59 $ & $ 73.21$  \\
    PCT  &   & $ 85.31 $ & $ 72.96 $ & $ 73.01 $ & $ 73.00 $ & $ 73.00 $ & $ 73.02 $ & $ 73.02 $ & $ 72.93 $ & $ 73.00 $ & $ 72.86 $ & $ 72.96 $ & $ 72.96 $ & $ 72.89 $ & $ 72.98 $ & $ 72.99 $ & $ 72.97$  \\
    CoLAP  &   & $ 84.37 $ & $ 72.00 $ & $ 77.23 $ & $ 75.77 $ & $ 75.93 $ & $ 78.74 $ & $ 77.23 $ & $ 69.18 $ & $ 75.67 $ & $ 65.17 $ & $ 71.62 $ & $ 72.46 $ & $ 65.41 $ & $ 73.81 $ & $ 73.95 $ & $ 73.16 $  \\

    \midrule 

    FT  &  \multirow{5}{*}{5}  & $ 84.07 $ & $ 72.04 $ & $ 76.53 $ & $ 74.67 $ & $ 74.50 $ & $ 76.81 $ & $ 76.07 $ & $ 66.73 $ & $ 74.64 $ & $ 61.42 $ & $ 70.68 $ & $ 71.15 $ & $ 63.37 $ & $ 72.81 $ & $ 72.90 $ & $ 71.74$  \\
    CA  &   & $ 84.07 $ & $ 72.33 $ & $ 77.57 $ & $ 75.66 $ & $ 75.69 $ & $ 78.2 $ & $ 77.16 $ & $ 69.57 $ & $ 75.83 $ & $ 63.85 $ & $ 71.66 $ & $ 72.35 $ & $ 66.81 $ & $ 74.06 $ & $ 74.26 $ & $ 73.21 $  \\
    PCT  &   & $ 85.31 $ & $ 72.38 $ & $ 77.03 $ & $ 74.61 $ & $ 75.21 $ & $ 78.36 $ & $ 77.14 $ & $ 67.88 $ & $ 75.76 $ & $ 59.31 $ & $ 72.26 $ & $ 71.38 $ & $ 65.03 $ & $ 74.49 $ & $ 74.46 $ & $ 72.52$  \\
    CoLAP w/ XRCL  &   & $ 84.37 $ & $ 71.71 $ & $ 78.54 $ & $ 76.47 $ & $ 76.44 $ & $ 79.28 $ & $ 77.56 $ & $ 69.83 $ & $ 75.82 $ & $ 64.25 $ & $ 73.19 $ & $ 72.17 $ & $ 65.63 $ & $ 74.84 $ & $ 74.05 $ & $ 73.56 $  \\
\bottomrule
    \end{tabularx}}
    \caption{Average accuracy per language on the XNLI dataset for XLM-R.}
    \label{tab:results_xlmr_xnli}
\end{table*}
\begin{table*}
\centering
{
\def\arraystretch{0.99}
\fontsize{5.5pt}{5.4pt}\selectfont
    \begin{tabularx}{\linewidth}{lc!{\vrule}>{\columncolor{gray!25}}Y*{14}{Y}!{\vrule}Y}
    \toprule
    \textbf{Model} &
    \textbf{K} &
    \textbf{en} &
    \textbf{ar} &
    \textbf{bg} &
    \textbf{de} &
    \textbf{el} &
    \textbf{es} &
    \textbf{fr} &
    \textbf{hi} &
    \textbf{ru} &
    \textbf{sw} &
    \textbf{th} &
    \textbf{tr} &
    \textbf{ur} &
    \textbf{vi} &
    \textbf{zh} &
    \textbf{Avg.} \\

    \midrule

    FT  &  \multirow{3}{*}{0}  & $87.76$ &  $57.74$ & $76.61$ & $77.29$ & $73.71$ & $81.34$ & $82.06$ & $61.22$ & $71.80$ & $62.38$ & $57.23$ & $71.62$ & $53.55$ & $73.17$ & $70.70$ & $69.32$ \\
    PCT  &   & $88.76$ & $68.18$ & $75.39$ & $78.84$ & $75.11$ & $80.56$ & $79.32$ & $69.80$ & $74.59$ & $62.50$ & $68.56$ & $67.96$ & $54.89$ & $72.53$ & $77.09$ & $71.81$ \\
    CoLAP  &   & $88.32$ & $74.51$ & $80.06$ & $78.32$ & $79.14$ & $84.15$ & $81.86$ & $68.26$ & $77.92$ & $64.15$ & $70.52$ & $70.64$ & $59.10$ & $76.19$ & $76.11$ & $74.35$ \\
    
    \midrule

    FT  &  \multirow{5}{*}{5}  & $87.76$ &  $61.57$ & $76.90$ & $77.75$ & $74.28$ & $81.31$ & $81.22$ & $64.12$ & $73.47$ & $63.72$ & $59.82$ & $71.43$ & $54.43$ & $73.99$ & $71.30$ & $70.38$ \\
    CA  &   & $87.76$ & $61.57$ & $76.90$ & $77.75$ & $74.28$ & $81.31$ & $81.22$ & $64.12$ & $73.47$ & $63.72$ & $59.82$ & $71.43$ & $54.43$ & $73.99$ & $71.30$ & $70.38$ \\
    PCT  &   & $88.76$ &  $69.76$ & $75.95$ & $79.52$ & $75.35$ & $80.84$ & $78.85$ & $70.43$ & $75.73$ & $62.98$ & $69.26$ & $69.01$ & $55.68$ & $73.39$ & $77.55$ & $72.45$ \\
    CoLAP w/ XRCL  &   & $88.32$ & $74.64$ & $79.72$ & $78.65$ & $79.00$ & $84.01$ & $81.92$ & $68.86$ & $77.85$ & $64.18$ & $70.71$ & $71.14$ & $60.67$ & $76.28$ & $76.61$ & $74.59$ \\
\bottomrule
    \end{tabularx}}
    \caption{Average accuracy per language on the XNLI dataset for Gemma 2 2B.}
    \label{tab:results_gemma_xnli}
\end{table*}
\begin{table*}
\centering
{
\def\arraystretch{0.99}
\fontsize{5.5pt}{5.4pt}\selectfont
    \begin{tabularx}{\linewidth}{lc!{\vrule}>{\columncolor{gray!25}}Y*{14}{Y}!{\vrule}Y}
    \toprule
    \textbf{Model} &
    \textbf{K} &
    \textbf{en} &
    \textbf{ar} &
    \textbf{bg} &
    \textbf{de} &
    \textbf{el} &
    \textbf{es} &
    \textbf{fr} &
    \textbf{hi} &
    \textbf{ru} &
    \textbf{sw} &
    \textbf{th} &
    \textbf{tr} &
    \textbf{ur} &
    \textbf{vi} &
    \textbf{zh} &
    \textbf{Avg.} \\

    \midrule

    FT  &  \multirow{3}{*}{0}  & $89.22$ &  $60.38$ & $68.00$ & $76.37$ & $63.29$ & $77.33$ & $75.51$ & $56.51$ & $68.10$ & $40.96$ & $52.77$ & $56.95$ & $49.38$ & $62.46$ & $59.66$ & $61.98$ \\
    PCT  &   & $90.32$ & $64.71$ & $80.66$ & $79.20$ & $58.50$ & $82.63$ & $81.32$ & $58.86$ & $79.42$ & $42.34$ & $60.90$ & $61.82$ & $51.90$ & $66.95$ & $78.26$ & $67.68$ \\
    CoLAP  &   & $90.12$ & $64.39$ & $77.49$ & $76.99$ & $61.84$ & $78.46$ & $77.94$ & $59.38$ & $76.13$ & $40.78$ & $62.02$ & $57.86$ & $52.34$ & $65.37$ & $76.29$ & $66.23$ \\
    
    \midrule

    FT  &  \multirow{5}{*}{5}  & $89.22$ & $61.37$ & $68.55$ & $76.07$ & $63.73$ & $77.65$ & $76.62$ & $57.99$ & $70.65$ & $40.99$ & $54.38$ & $57.66$ & $50.13$ & $63.15$ & $61.72$ & $62.90$ \\
    CA  &   & $89.22$ & $61.37$ & $68.55$ & $76.07$ & $63.73$ & $77.65$ & $76.62$ & $57.99$ & $70.65$ & $40.99$ & $54.38$ & $57.66$ & $50.13$ & $63.15$ & $61.72$ & $62.90$ \\
    PCT  &   & $90.32$ & $65.66$ & $80.61$ & $79.29$ & $58.03$ & $82.75$ & $81.44$ & $58.85$ & $79.50$ & $42.72$ & $61.48$ & $61.98$ & $53.34$ & $67.10$ & $78.37$ & $67.94$ \\
    CoLAP w/ XRCL  &   & $90.12$ & $66.43$ & $78.79$ & $78.72$ & $64.01$ & $79.34$ & $78.71$ & $62.12$ & $77.97$ & $41.56$ & $64.15$ & $59.36$ & $54.47$ & $65.41$ & $76.54$ & $67.68$ \\
\bottomrule
    \end{tabularx}}
    \caption{Average accuracy per language on the XNLI dataset for Mistral 7B.}
    \label{tab:results_mistral_xnli}
\end{table*}

\begin{table*}
\centering
{
\def\arraystretch{0.99}
\fontsize{5.5pt}{5.4pt}\selectfont
    \begin{tabularx}{\linewidth}{lc!{\vrule}*{10}{Y}!{\vrule}Y}
    \toprule
    \textbf{Model} &
    \textbf{K} &
    \textbf{aym} &
    \textbf{bzd} &
    \textbf{cni} &
    \textbf{gn} &
    \textbf{hch} &
    \textbf{nah} &
    \textbf{oto} &
    \textbf{quy} &
    \textbf{shp} &
    \textbf{tar} &
    \textbf{Avg.} \\

    \midrule

    FT  &  \multirow{3}{*}{0}  & $ 36.80 $ & $ 37.60 $ & $ 36.27 $ & $ 38.13 $ & $ 37.20 $ & $ 37.94 $ & $ 35.96 $ & $ 38.27 $ & $ 35.73 $ & $ 35.20 $ & $ 36.91 $  \\
    PCT  &   & $ 38.42 $ & $ 38.16 $ & $ 38.02 $ & $ 37.89 $ & $ 36.96 $ & $ 40.81 $ & $ 38.93 $ & $ 36.82 $ & $ 36.29 $ & $ 37.49 $ & $ 37.98 $  \\
    CoLAP &   & $ 35.47 $ & $ 35.47 $ & $ 34.93 $ & $ 36.53 $ & $ 34.80 $ & $ 37.67 $ & $ 35.83 $ & $ 36.53 $ & $ 35.07 $ & $ 35.60 $ & $ 35.79 $  \\

    \midrule 

    FT  &  \multirow{5}{*}{5}  & $ 39.17 $ & $ 40.53 $ & $ 40.37 $ & $ 38.93 $ & $ 38.07 $ & $ 40.07 $ & $ 40.17 $ & $ 40.30 $ & $ 40.63 $ & $ 37.47 $ & $ 39.57 $  \\
    CA  &   & $ 37.60 $ & $ 38.77 $ & $ 39.27 $ & $ 37.27 $ & $ 36.57 $ & $ 39.60 $ & $ 38.34 $ & $ 36.77 $ & $ 38.83 $ & $ 36.43 $ & $ 37.95 $  \\
    PCT  &   & $ 37.64 $ & $ 40.87 $ & $ 38.58 $ & $ 39.64 $ & $ 38.87 $ & $ 39.48 $ & $ 40.02 $ & $ 37.62 $ & $ 39.27 $ & $ 39.14 $ & $ 39.11 $  \\
    CoLAP w/ XRCL  &   & $ 38.77 $ & $ 40.69 $ & $ 39.28 $ & $ 41.57 $ & $ 38.88 $ & $ 40.11 $ & $ 40.24 $ & $ 40.29 $ & $ 39.68 $ & $ 40.59 $ & $ 40.01 $  \\
\bottomrule
    \end{tabularx}}
    \caption{Average accuracy per language on the AmNLI dataset for XLM-R.}
    \label{tab:results_xlmr_anli}
\end{table*}
\begin{table*}
\centering
{
\def\arraystretch{0.99}
\fontsize{5.5pt}{5.4pt}\selectfont
    \begin{tabularx}{\linewidth}{lc!{\vrule}*{10}{Y}!{\vrule}Y}
    \toprule
    \textbf{Model} &
    \textbf{K} &
    \textbf{aym} &
    \textbf{bzd} &
    \textbf{cni} &
    \textbf{gn} &
    \textbf{hch} &
    \textbf{nah} &
    \textbf{oto} &
    \textbf{quy} &
    \textbf{shp} &
    \textbf{tar} &
    \textbf{Avg.} \\

    \midrule

    FT  &  \multirow{3}{*}{0}  &  $42.00$ & $40.93$ & $40.67$ & $42.93$ & $38.27$ & $43.63$ & $38.37$ & $40.67$ & $41.87$ & $40.53$ & $40.99$ \\
    PCT  &   & $39.73$ & $42.53$ & $39.07$ & $46.53$ & $36.53$ & $44.99$ & $41.58$ & $41.07$ & $42.13$ & $38.53$ & $41.27$ \\
    CoLAP  &  & $42.00$ & $44.40$ & $42.13$ & $48.00$ & $38.80$ & $47.97$ & $40.37$ & $42.27$ & $42.00$ & $39.33$ & $42.73$ \\

    \midrule

    FT  &  \multirow{5}{*}{5}  & $42.24$ & $40.53$ & $40.16$ & $45.09$ & $38.32$ & $44.04$ & $38.66$ & $41.41$ & $42.64$ & $40.96$ & $41.41$ \\
    CA  &   & $42.24$ & $40.53$ & $40.16$ & $45.09$ & $38.32$ & $44.04$ & $38.66$ & $41.41$ & $42.64$ & $40.96$ & $41.41$ \\
    PCT  &   & $42.80$ & $43.11$ & $40.04$ & $47.60$ & $36.00$ & $45.44$ & $40.78$ & $41.07$ & $43.24$ & $39.87$ & $41.99$ \\
    CoLAP w/ XRCL  &  &$41.79$ & $43.71$ & $42.21$ & $46.45$ & $39.97$ & $47.67$ & $40.78$ & $43.47$ & $42.99$ & $39.76$ & $42.88$ \\
\bottomrule
    \end{tabularx}}
    \caption{Average accuracy per language on the AmNLI dataset for Gemma 2 2B.}
    \label{tab:results_gemma_anli}
\end{table*}
\begin{table*}
\centering
{
\def\arraystretch{0.99}
\fontsize{5.5pt}{5.4pt}\selectfont
    \begin{tabularx}{\linewidth}{lc!{\vrule}*{10}{Y}!{\vrule}Y}
    \toprule
    \textbf{Model} &
    \textbf{K} &
    \textbf{aym} &
    \textbf{bzd} &
    \textbf{cni} &
    \textbf{gn} &
    \textbf{hch} &
    \textbf{nah} &
    \textbf{oto} &
    \textbf{quy} &
    \textbf{shp} &
    \textbf{tar} &
    \textbf{Avg.} \\

    \midrule

    FT  &  \multirow{3}{*}{0}  & $37.47$ & $37.60$ & $38.40$ & $37.07$ & $36.93$ & $42.68$ & $36.10$ & $37.07$ & $37.87$ & $34.80$ & $37.60$ \\
    PCT  &   & $36.93$ & $38.00$ & $36.13$ & $38.40$ & $35.47$ & $39.57$ & $36.90$ & $37.33$ & $38.67$ & $36.27$ & $37.37$ \\
    CoLAP  &  & $37.33$ & $38.00$ & $36.27$ & $37.47$ & $36.00$ & $39.84$ & $37.30$ & $38.40$ & $37.47$ & $33.73$ & $37.18$ \\

    \midrule

    FT  &  \multirow{5}{*}{5}  & $37.68$ & $36.51$ & $38.32$ & $36.40$ & $35.47$ & $42.09$ & $36.23$ & $38.43$ & $39.63$ & $35.04$ & $37.58$ \\
    CA  &   &$37.68$ & $36.51$ & $38.32$ & $36.40$ & $35.47$ & $42.09$ & $36.23$ & $38.43$ & $39.63$ & $35.04$ & $37.58$ \\
    PCT  &   & $39.80$ & $38.53$ & $40.07$ & $39.93$ & $38.20$ & $41.60$ & $38.24$ & $39.87$ & $41.87$ & $36.53$ & $39.46$ \\
    CoLAP w/ XRCL  & & $40.93$ & $39.49$ & $38.16$ & $41.49$ & $37.15$ & $41.52$ & $38.13$ & $40.00$ & $40.08$ & $35.17$ & $39.21$ \\
\bottomrule
    \end{tabularx}}
    \caption{Average accuracy per language on the AmNLI dataset for Mistral 7B.}
    \label{tab:results_mistral_anli}
\end{table*}

\begin{table*}
\centering
{
\def\arraystretch{0.99}
\fontsize{5.5pt}{5.4pt}\selectfont
    \begin{tabularx}{\linewidth}{lc!{\vrule}>{\columncolor{gray!25}}Y*{12}{Y}!{\vrule}Y}
    \toprule
    \textbf{Model} &
    \textbf{K} &
    \textbf{en} &
    \textbf{ar} &
    \textbf{de} &
    \textbf{es} &
    \textbf{fi} &
    \textbf{fr} &
    \textbf{hi} &
    \textbf{hu} &
    \textbf{ja} &
    \textbf{pl} &
    \textbf{ru} &
    \textbf{tr} &
    \textbf{zh} &
    \textbf{Avg.} \\

    \midrule

    FT  &  \multirow{3}{*}{0}  & $ 85.18 $ & $ 23.20 $ & $ 26.32 $ & $ 30.96 $ & $ 25.61 $ & $ 34.29 $ & $ 22.8 $ & $ 24.80 $ & $ 12.61 $ & $ 23.60 $ & $ 29.58 $ & $ 17.20 $ & $ 23.90 $ & $ 24.57$  \\
    PCT  &   & $ 87.38 $ & $ 40.40 $ & $ 49.39 $ & $ 54.39 $ & $ 50.81 $ & $ 53.88 $ & $ 38.00 $ & $ 46.00 $ & $ 41.44 $ & $ 50.00 $ & $ 52.50 $ & $ 33.60 $ & $ 40.00 $ & $ 45.87 $  \\
    CoLAP  &   & $ 87.94 $ & $ 55.60 $ & $ 48.99 $ & $ 57.74 $ & $ 50.41 $ & $ 60.41 $ & $ 47.60 $ & $ 46.40 $ & $ 49.55 $ & $ 49.20 $ & $ 60.83 $ & $ 40.00 $ & $ 52.20 $ & $ 51.58 $  \\

    \midrule 

    FT  &  \multirow{5}{*}{5}  & $ 85.18 $ & $ 42.88 $ & $ 44.37 $ & $ 45.44 $ & $ 43.50 $ & $ 45.55 $ & $ 40.16 $ & $ 41.52 $ & $ 35.41 $ & $ 44.08 $ & $ 45.42 $ & $ 39.20 $ & $ 44.29 $ & $ 42.65 $  \\
    CA  &   & $ 85.18 $ & $ 37.76 $ & $ 41.13 $ & $ 45.10 $ & $ 40.41 $ & $ 45.06 $ & $ 36.32 $ & $ 34.88 $ & $ 31.44 $ & $ 38.24 $ & $ 43.08 $ & $ 32.72 $ & $ 41.07 $ & $ 38.93$  \\
    PCT  &   & $ 87.38 $ & $ 69.80 $ & $ 71.56 $ & $ 73.01 $ & $ 69.61 $ & $ 73.27 $ & $ 67.10 $ & $ 71.60 $ & $ 64.64 $ & $ 66.80 $ & $ 71.56 $ & $ 65.50 $ & $ 63.66 $ & $ 69.01 $  \\
    CoLAP w/ XRCL  &   & $ 87.94 $ & $ 71.60 $ & $ 72.15 $ & $ 73.05 $ & $ 70.00 $ & $ 71.76 $ & $ 66.24 $ & $ 70.64 $ & $ 63.96 $ & $ 69.36 $ & $ 70.25 $ & $ 66.96 $ & $ 65.17 $ & $ 69.26 $  \\
\bottomrule
    \end{tabularx}}
    \caption{Average accuracy per language on the MultiTACRED dataset for XLM-R.}
    \label{tab:results_xlmr_tacred}
\end{table*}
\begin{table*}
\centering
{
\def\arraystretch{0.99}
\fontsize{5.5pt}{5.4pt}\selectfont
    \begin{tabularx}{\linewidth}{lc!{\vrule}>{\columncolor{gray!25}}Y*{12}{Y}!{\vrule}Y}
    \toprule
    \textbf{Model} &
    \textbf{K} &
    \textbf{en} &
    \textbf{ar} &
    \textbf{de} &
    \textbf{es} &
    \textbf{fi} &
    \textbf{fr} &
    \textbf{hi} &
    \textbf{hu} &
    \textbf{ja} &
    \textbf{pl} &
    \textbf{ru} &
    \textbf{tr} &
    \textbf{zh} &
    \textbf{Avg.} \\

    \midrule

    FT  &  \multirow{3}{*}{0}  & $85.03$ & $10.91$ & $21.68$ & $21.85$ & $13.72$ & $15.97$ & $9.86$ & $18.93$ & $4.91$ & $16.64$ & $14.88$ & $15.08$ & $10.20$ & $14.55$ \\
    PCT  &   & $87.10$ & $23.34$ & $35.84$ & $46.26$ & $29.13$ & $46.80$ & $34.48$ & $36.49$ & $22.01$ & $34.32$ & $42.13$ & $29.27$ & $29.80$ & $34.16$ \\
    CoLAP  &   & $87.05$ &$35.74$ & $37.20$ & $50.41$ & $36.44$ & $44.84$ & $24.94$ & $31.35$ & $17.21$ & $27.76$ & $35.63$ & $21.49$ & $26.70$ & $32.48$ \\

    \midrule

    FT  &  \multirow{5}{*}{5}  & $85.03$ & $11.71$ & $26.04$ & $24.12$ & $18.89$ & $19.91$ & $13.27$ & $20.49$ & $6.58$ & $21.72$ & $16.77$ & $18.33$ & $12.49$ & $17.53$ \\
    CA  &   & $85.03$ &  $11.71$ & $26.04$ & $24.12$ & $18.89$ & $19.91$ & $13.27$ & $20.49$ & $6.58$ & $21.72$ & $16.77$ & $18.33$ & $12.49$ & $17.53$ \\
    PCT  &   & $87.10$ & $33.04$ & $48.64$ & $56.37$ & $42.20$ & $57.74$ & $42.98$ & $44.42$ & $30.04$ & $44.40$ & $48.93$ & $39.21$ & $39.99$ & $44.00$ \\
    CoLAP w/ XRCL  &   & $87.05$ & $49.00$ & $44.40$ & $66.53$ & $46.71$ & $57.06$ & $33.44$ & $41.30$ & $28.75$ & $40.72$ & $45.18$ & $32.80$ & $35.75$ & $43.47$ \\
\bottomrule
    \end{tabularx}}
    \caption{Average accuracy per language on the MultiTACRED dataset for Gemma 2 2B.}
    \label{tab:results_gemma_tacred}
\end{table*}
\begin{table*}
\centering
{
\def\arraystretch{0.99}
\fontsize{5.5pt}{5.4pt}\selectfont
    \begin{tabularx}{\linewidth}{lc!{\vrule}>{\columncolor{gray!25}}Y*{12}{Y}!{\vrule}Y}
    \toprule
    \textbf{Model} &
    \textbf{K} &
    \textbf{en} &
    \textbf{ar} &
    \textbf{de} &
    \textbf{es} &
    \textbf{fi} &
    \textbf{fr} &
    \textbf{hi} &
    \textbf{hu} &
    \textbf{ja} &
    \textbf{pl} &
    \textbf{ru} &
    \textbf{tr} &
    \textbf{zh} &
    \textbf{Avg.} \\

    \midrule

    FT  &  \multirow{3}{*}{0}  & $85.08$ & $6.26$ & $18.00$ & $13.60$ & $9.30$ & $16.20$ & $5.29$ & $12.51$ & $5.99$ & $12.32$ & $15.39$ & $9.70$ & $6.78$ & $10.95$ \\
    PCT  &   & $87.68$ & $5.13$ & $48.40$ & $53.42$ & $30.26$ & $50.51$ & $8.74$ & $39.30$ & $23.47$ & $42.96$ & $42.82$ & $25.98$ & $37.60$ & $34.05$ \\
    CoLAP  &   & $85.68$ & $19.00$ & $53.44$ & $50.77$ & $39.56$ & $55.27$ & $18.85$ & $52.13$ & $29.41$ & $43.52$ & $33.22$ & $32.24$ & $37.34$ & $38.73$ \\

    \midrule

    FT  &  \multirow{5}{*}{5}  & $85.08$ & $8.82$ & $21.48$ & $20.93$ & $12.36$ & $18.97$ & $7.06$ & $13.88$ & $6.67$ & $14.76$ & $16.40$ & $10.51$ & $7.02$ & $13.24$ \\
    CA  &   & $85.08$ & $8.82$ & $21.48$ & $20.93$ & $12.36$ & $18.97$ & $7.06$ & $13.88$ & $6.67$ & $14.76$ & $16.40$ & $10.51$ & $7.02$ & $13.24$ \\
    PCT  &   & $87.68$ & $12.19$ & $58.56$ & $59.87$ & $42.05$ & $57.86$ & $15.72$ & $47.96$ & $40.33$ & $53.44$ & $49.41$ & $35.84$ & $46.92$ & $43.35$ \\
    CoLAP w/ XRCL  &   & $85.68$ & $19.00$ & $53.44$ & $50.77$ & $39.56$ & $55.27$ & $18.85$ & $52.13$ & $29.41$ & $43.52$ & $33.22$ & $32.24$ & $37.34$ & $38.73$ \\
\bottomrule
    \end{tabularx}}
    \caption{Average accuracy per language on the MultiTACRED dataset for Mistral 7B.}
    \label{tab:results_mistral_tacred}
\end{table*}

\begin{table*}
\centering
\scalebox{0.8}{
    \begin{tabular}{llp{10cm}}
    \hline
    \textbf{Dataset} &
    \textbf{Label} &
    \textbf{Exemplar} \\
    \hline

    XNLI & Entailment & Premise: ``You look like f--ing hell , Brando , the Star reports he said , and he advised the actor to lose maybe a hundred pounds , pallie .'' Hypothesis: ``Brando looks like f--ing hell , the Star reported him as saying .'' \\
    & Neutral & Premise: ``3 ) Their many enemies want to legislate them out of existence .'' Hypothesis: ``The Democrats have enemies that want to legislate them out of existence .'' \\
    & Neutral & Premise: ``Only three now , sir .'' Hypothesis: ``3 pm now , sir .'' \\
    & Contradiction & Premise: ``That would certainly help i 'm sure.'' Hypothesis: ``No , that does not help me .'' \\
    & Contradiction & Premise: ``They ate them , uncle .'' Hypothesis: ``The people were unharmed .'' \\

    \hline

    AmNLI & Entailment & Premise: ``That was my messed up.'' Hypothesis: ``I made a mistake.'' \\
    & Neutral & Premise: ``They asked a few questions and I answered them and they said, Get your baggage and leave there immediately, and come to the address you were supposed to when you arrived in Washington.'' Hypothesis: ``They told me to pick up by white suitcase.'' \\
\hline
    \end{tabular}}
    \caption{Exemplars with highest similarity scores for XNLI, AmNLI, and MultiTACRED.}
    \label{tab:exemplars}
\end{table*}

\end{document}